\documentclass[sigconf]{acmart}
\AtBeginDocument{%
  }

\copyrightyear{2026}
\acmYear{2026}
\setcopyright{cc}
\setcctype{by}
\acmConference[MM '26]{Proceedings of the 34th ACM International Conference on Multimedia}{November 10--14, 2026}{Rio de Janeiro, Brazil}
\acmBooktitle{Proceedings of the 34th ACM International Conference on Multimedia (MM '26), November 10--14, 2026, Rio de Janeiro, Brazil}
\acmDOI{10.1145/3767308.3836423}
\acmISBN{979-8-4007-2213-4/2026/11}

\usepackage{cleveref}
\crefname{figure}{Fig.}{Fig.}
\crefname{table}{Tab.}{Tab.}
\crefname{section}{Sec.}{Sec.}
\usepackage{multirow}

\begin{document}

\title{Estimating Accurate Hand Pose in Camera Space with Vision Transformer}


\author{Kaiwen Ren}
\authornote{Both authors contributed equally to this research.}
\email{renkaiwen23s@ict.ac.cn}
\orcid{0009-0002-4372-3463}
\author{Yiran Jiang}
\authornotemark[1]
\email{jiangyiran25s@ict.ac.cn}
\orcid{0009-0007-7539-8542}
\affiliation{%
  \institution{Institute of Computing Technology, Chinese Academy of Sciences}
  \institution{University of Chinese Academy of Sciences}
  \city{Beijing}
  \country{China}
}

\author{Yongjing Ye}
\orcid{0000-0002-1027-3382}
\affiliation{%
  \institution{Institute of Computing Technology, Chinese Academy of Sciences}
  \city{Beijing}
  \country{China}}
\email{yeyongjing@ict.ac.cn}

\author{Shihong Xia}
\correspondingauthor
\authornote{Corresponding author.}
\orcid{0000-0002-7228-9646}
\affiliation{%
  \institution{Institute of Computing Technology, Chinese Academy of Sciences}
  \institution{University of Chinese Academy of Sciences}
  \city{Beijing}
  \country{China}
}
\email{xsh@ict.ac.cn}

\renewcommand{\shortauthors}{Kaiwen Ren, Yiran Jiang, Yongjing Ye and Shihong Xia}

\begin{abstract}
  Monocular RGB-based hand pose estimation has emerged as a critical research frontier in computer vision. The local hand pose estimation methods predict hand poses relative to the wrist, while global hand pose estimation also requires estimating the wrist's position in the camera coordinate system. However, this camera-space estimation confronts two fundamental challenges: (1) depth ambiguity in monocular settings, and (2) the coupling effect of hand local poses and global wrist positions in the perspective projections. In particular, this coupling reflects that the projections are jointly determined by local hand poses, wrist positions, and camera intrinsics. To overcome these challenges, our framework proposes two key innovations: Transformation-Isomorphism Supervision for hand-depth information extraction and Perspective Information Embedding for resolving above coupling effect of local pose and wrist position, both integrated within the mainstream encoder-decoder architecture. Besides, we propose a novel framerate-aware multi-dataset training strategy for sequential pose refinement. Our fully integrated approach achieves at most 37.1\% superiority in CS-MJE over SOTA on HO3D. Project page: \url{https://github.com/Mine268/CS-ViT}.
\end{abstract}


\begin{teaserfigure}
  \includegraphics[width=\textwidth]{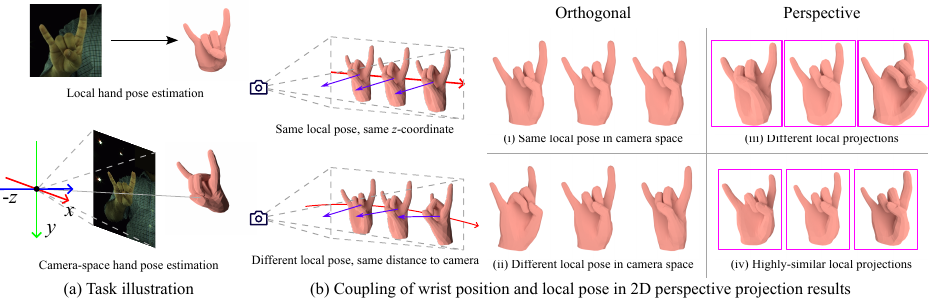}
  \caption{Task illustration and challenges. (a) compares local versus camera-space pose estimation, where local estimation predicts joint-relative articulation while camera-space estimation recovers under camera's coordinate. (b) demonstrates the perspective coupling effect. The perspective projection can distorts the projection of same local pose as (i) and (iii). It can also compensate for difference in local pose and yield the same projection results as in (ii) and (iv).}
  \label{fig:task-def-and-challenge}
\end{teaserfigure}


\maketitle


\section{Introduction}
Accurate monocular RGB-based hand pose estimation plays a pivotal role for VR/AR systems and Human-Computer Interaction (HCI) applications, by facilitating seamless interaction with virtual objects without additional sensors. As shown in Fig \ref{fig:task-def-and-challenge}(a), \textbf{local hand pose estimation} methods aim to regress wrist-relative hand poses, whereas \textbf{camera-space hand pose estimation} requires localizing the wrist joint in camera coordinates simultaneously. Camera-space hand pose estimation could provide more comprehensive hand information, therefore becoming the primary focus in our work.

Camera-space hand pose estimation faces two major challenges. First, inherent depth ambiguity in monocular views arises when objects of varying sizes at different distances yield identical projections, which is a common limitation in hand pose estimation. Local hand pose estimation methods try to estimate accurate hand size by introducing prior information \cite{MANO:SIGGRAPHASIA:2017, spurr_peclr_2022}. However, predicting camera-space hand pose not only requires accurate estimation of hand size, but also depth inference based on that size estimation \cite{handdgp:leonardis_computer_2025,nvf:huang_neural_2023}. This imposes stricter requirements on the model's ability to perceive depth.

The second challenge is the irreducible coupling result of wrist positions and local hand poses under perspective projection, we claimed it as ``perspective coupling effect''. As shown in \cref{fig:task-def-and-challenge}(b), orthographic projection precludes perspective effects, thereby rendering the influences of local pose and root joint position mutually independent in the projected result. Consequently, the root joint's $x,y$ coordinates and local orientation can be directly inferred from the image, as illustrated in (i) and (ii). In contrast, perspective projection entangles root position and local pose, both contributing to the 2D hand orientation in the projection. This coupling makes it impossible to determine the relative contribution of each factor from the image alone, thereby complicating pose estimation. An extreme example is shown in (iv): three distinct combinations of local pose and root joint position produce identical 2D hand projections.

Perspective effects are therefore indispensable in camera-space pose estimation. Accurate recovery necessitates explicit modeling of camera intrinsics and perspective geometry. Most existing approaches\cite{blur_oh2023recovering3dhandmesh,potamias2024wilor,h2onet:Xu_2023_CVPR,deformer:Fu_2023_ICCV,10655481} either overlook this requirement entirely or rely on virtual intrinsic assumptions; even the latter typically restrict such assumptions to auxiliary 2D projection loss computation, thereby introducing systematic errors into camera-space estimation. For full theoretical analysis please refer to Appendix.

\begin{figure}[!t]
    \centering
    \includegraphics[width=0.7\linewidth]{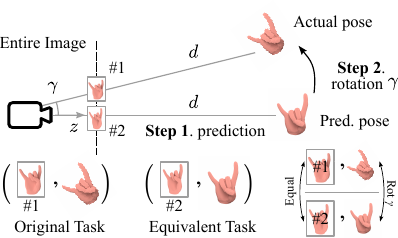}
    \caption{The illustration of our intuitive baseline. The cropping image (\#1) is treated as if it is taken by pointing camera directly to the hand (\#2). Then the predicted dummy pose is rotated to match the actual position of the cropping, giving prediction of actual pose.}
    \label{fig:our-baseline}
\end{figure}

To address the aforementioned challenges, the model must (1) sensitively incorporate hand priors to extract depth information from hand images and (2) precisely perceive perspective information of the hand projection. For the first requirement, inspired by \cite{10.2312:pg.20251270}, we propose Transformation-Isomorphism Supervision (TIS), which aligns the scale and depth information through ``rotation'' \& ``scaling'' in the latent space to enhance the perception of them for the backbone. For the second requirement, the Perspective Information Embedding (PIE) mechanism is inserted, enabling the ability of perspective information awareness by decorating the image latent patches with perspective vector. The significance of PIE is demonstrated through our carefully designed baseline (\cref{fig:our-baseline}), which builds upon the observations in \cref{fig:task-def-and-challenge}(b,iv) by first predicting a canonical dummy pose and then applying view-specific transformations to match the input image perspective. In addition, we propose framerate-aware multi-dataset approach to train our model on multiple datasets with various capture configurations, improving training data adaptability and pose estimation accuracy. In conclusion, our contributions are summarized as follows:

\begin{enumerate}
    \item The concept of perspective coupling effect is formalized, with qualitative visualization and theoretical analysis of its impact on hand pose estimation.
    \item Transformation-Isomorphism Supervision (TIS): A training component that improves the ability to perceive depth information and alleviates the ambiguity of depth.
    \item Perspective Information Embedding (PIE): A mechanism that helps the model to perceive the perspective information and to model the coupling of perspective on hand pose in the latent space.
    \item Framerate-Aware Multi-Dataset Training (MDT): This mechanism supports joint training across datasets with varying framerate, improving the performance of our pose estimation network.
\end{enumerate}

By integrating the TIS and PIE mechanisms, we achieve accurate camera-space hand pose estimation, demonstrating 7.8\% improvement over state-of-the-art methods \cite{handdgp:leonardis_computer_2025} on the HO3D dataset. This performance gain further expands to around 37.1\% when incorporating the framerate-aware multi-dataset training mechanism.


\section{Related Works}
\subsection{Local hand pose estimation}
In recent years, research has been devoted to hand pose estimation, especially predicting local hand poses from monocular RGB-based image or sequence \cite{mmpose2020,Lugaresi2019MediaPipeAF,xu2022vitpose,xu2022vitpose+,Hampali_2022_CVPR_Kypt_Trans,10.1109/TPAMI.2023.3247907,10655481,10.1007/978-3-031-20068-7_22,moon_bringing_2023,blur_oh2023recovering3dhandmesh,ego_Prakash2024Hands,lin_pre-training_2024,spurr_peclr_2022,Zhao_2025_CVPR}.

Among various architectures, transformer-based models outperform others for its high accuracy and robustness \cite{Potamias_2025_CVPR,10655481} against other architectures like CNN \cite{zhou2024simple,Lugaresi2019MediaPipeAF,Hampali_2020_CVPR} or graph-CNN \cite{ge2019handshapepose}. It has been proven that scaling laws \cite{kaplan2020scalinglawsneurallanguage} still roughly hold in this field.

\begin{figure*}[!t]
	\centering
	\includegraphics[width=0.95\linewidth]{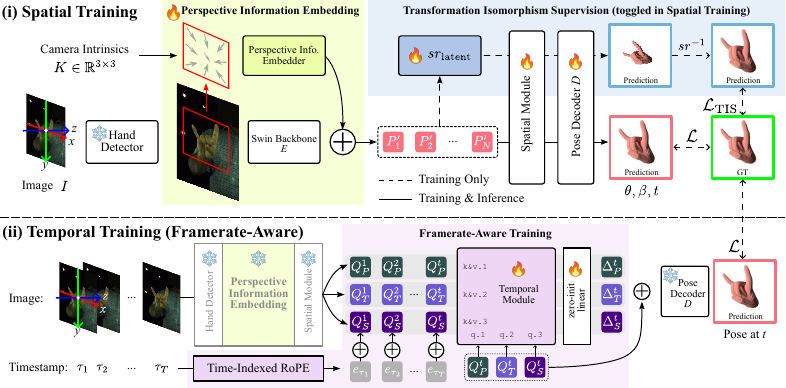}
	\caption{Method illustration. (i) In single-frame training, an off-the-shelf detector localizes the hand, then \textbf{Perspective Information Embedding} (PIE) enhances latent feature with camera intrinsics and detection boxes. The spatial module extracts pose features, which are mapped to pose parameter by the Pose Decoder. \textbf{Transformation Isomorphism Supervision} (TIS) transforms patches and compares decoded poses with GT, imposing depth information into latent space. (ii) Temporal Module is added after the spatial module to exploit the temporal cues. We design the zero-init linear layer for training efficiency. The TIS mechanism is omitted in temporal training stage.}
	\label{fig:method-st}
\end{figure*}

Specialized designs have also been proposed to resolve the inherent challenges of hand pose estimation: occlusion, self-occlusion and motion blur. H2ONet \cite{h2onet:Xu_2023_CVPR} proposed learning occlusion information for hand-object interaction via dual-branch encoder. Deformer \cite{deformer:Fu_2023_ICCV} proposed a dynamic fusion transformer for temporal pose modeling to address motion blur. By incorporating vision transformer with large scale of data, HaMeR\cite{10655481} achieves impressive accuracy and robustness in local hand pose estimation. Based on that, WiLoR \cite{potamias2024wilor} advances multi-hand localization and estimation with the novel image-aligned multi-scale sampling and a dataset including 2M+ in-the-wild images.

Though lots of works with various architectures emerged to tackle the challenges in this field, few works provide solutions for global localization of hand in the camera space, leaving a substantial space for our work.

\subsection{Hand pose estimation in camera space}
Camera-space and global hand pose estimation has drawn the attention of cutting-edge research. This distinct task focuses on estimating the pose of hand, including the position relative to camera. Recent studies \cite{yu2024dynhamr,zhang2025hawor} integrate local hand pose estimation methods\cite{10655481,Potamias_2025_CVPR} with SLAM-based camera tracking, enabling continuous hand pose tracking in world coordinates. We categorize them as global hand pose estimation.

For camera coordinate-based estimation, recent approaches address this challenge through novel paradigms. Huang et al. \cite{nvf:huang_neural_2023} propose a Neural Voting Field (NVF) that projects 2D features into 3D voxels to predict joint-wise voting preferences, thereby locating hand joints in camera coordinates.  This projection mechanism circumvents coupling challenges via voxel representation and reprojection. However, their approach predicts the joint's position instead of rotation due to the voxel restriction, limiting the applicability. Valassakis et al. \cite{handdgp:leonardis_computer_2025} introduce HandDGP, which established  differentiable PnP solutions for derivable 2D-3D lifting. Their PnP algorithm bridges the ``local to camera-space'' estimation in learnable approach. However, they neglect the perspective coupling effect, resulting in accuracy degradation compared to our method, as shown in our experiment at \cref{tab:ho3d}.

\subsection{Representation Learning for hand pose estimation}

Some studies employ representation learning to enhance pose estimation methods. The goal of representation learning is to learn representations beneficial for downstream tasks, thereby improving both the performance of training and inference. As categorized by \cite{singh_explainable_2024}, representation learning comprises two paradigms: MIM-based methods reconstruct images to learn representations \cite{MaskedAutoencoders2021}, while contrastive-based methods leverage positive-negative sample pairs for loss-driven representation learning \cite{pmlr-v119-chen20j}.

Most representation learning frameworks for hand pose estimation employ contrastive-based methods. For example, SiMHand \cite{lin2025simhand} constructs positive sample pairs by identifying images with similar poses via 2D pose detectors to enhance robustness against texture/lighting factors. We argue this is essentially knowledge distillation from 2D pose estimators. PeCLR \cite{spurr_peclr_2022} adapts SimCLR \cite{chen2020simple} framework by introducing geometric transformations to feature maps. This method directly applies a geometric inverse transformation in the image space to 2D feature maps, thereby directly constraining strong geometric priors in the latent space. We express concern about the practice of directly injecting intuitive constraints into the latent space. Similarly, \cite{10.2312:pg.20251270} also argues that such an approach can be further improved. Inspired by this, we propose Transformation-Isomorphic Supervision (TIS) to learn pose-enriched representations through parametric mappings isomorphic to geometry transformations across image and 3D domains.


\begin{figure}[!t]
    \centering
    \includegraphics[width=0.8\linewidth]{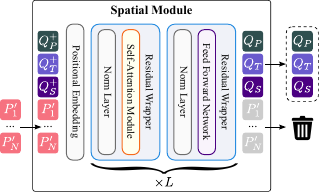}
    \caption{The spatial module extracts pose information from latent patches with query token $Q_P^+, Q_T^+, Q_S^+$. The latent patches $P_1',\cdots,P_N'$ are discarded  at the final layer.}
    \label{fig:spatial-module}
\end{figure}

\section{Method}

\subsection{Overview of Our Pipeline}

The spatial and temporal training phases are illustrated in \cref{fig:method-st}. During the spatial training phase, an off-the-shelf hand detector is adopted to obtain hand bounding boxes from input images \( I \in \mathbb{I} \). Subsequently, we utilize a Swin \cite{liu2021swinv2} backbone \( E: \mathbb{I} \to \mathbb{R}^{N \times d} \) to encode the cropped hand regions into latent patches \( \{P_i\in\mathbb R^d\}_{i=1}^N \).

The \textit{Perspective Information Embedding} mechanism leverages both the hand bounding box information and camera intrinsics to refine the latent patches into \( \{P'_i\}_{i=1}^N \), which now incorporate pose information in the camera space. \( \{P'_i\}_{i=1}^N \) are then processed by the Spatial Module (see \cref{fig:spatial-module}) to extract the tokens for pose feature \( Q_P, Q_T,Q_S\in\mathbb R^d \). Finally, a Pose Decoder \( D: \mathbb{R}^{3 \times d} \to \mathbb{R}^{(16\times 3)+10+3} \) decodes these features into the global hand pose \( (\theta,\beta,t) \) for joint rotation, shape, and wrist translation supervision respectively. In addition to supervising $\theta,\beta$ and $t$, we introduce a novel \textit{Transformation Isomorphism Supervision} (TIS) mechanism. Specifically, we first apply learnable ``scale-rotation'' transformation $sr_\text{latent}$ to \( \{P'_i\}_{i=1}^N \) and decode it into pose $(\theta',\beta',t')$ parameters using the same Pose Decoder. Then $(\theta',\beta',t')$ is inversely transformed in 3D space and is also supervised against GT.

During the temporal training phase, as illustrated in \cref{fig:method-st}(ii), we freeze the backbone and Spatial Module. We exclusively train the newly added Temporal Module, which is inserted after the Spatial Module. The Temporal Module takes as input the per-frame pose features $\{Q_P^i\}_{i=1}^t$, $\{Q_T^i\}_{i=1}^t$, and $\{Q_S^i\}_{i=1}^t$, and performs temporal information fusion to produce the refined pose features of the last frame, namely $Q_P^t$, $Q_T^t$, and $Q_S^t$. These fused features are then decoded into pose parameters using the Pose Decoder from the previous spatial training phase.

\subsection{Perspective Information Embedding}
\label{sec:method-pie}

The \cref{fig:task-def-and-challenge} demonstrates the coupling effect of wrist position and local hand pose: the projection of hand is the combination of hand pose and perspective information. Thus, we need a component to model the mechanism of this coupling process. Let us informally denote the 3D hand pose as $X$, the perspective information as $K$, and the 2D hand projection as $Y$. The projection process can be expressed as $Y=\Pi(X,K)$, where the $\Pi$ is the perspective projection operator. For precise 3D pose recovery $X$, the 2D hand projection $Y$ is inherently ambiguous without incorporating $K$. The recovery process can be expressed as $X=\Pi^{-1}(Y,K)$, with the symbol $\Pi^{-1}$ being the recovery process of hand pose estimation conditioned on $K$. The $K$ is self-explained but the information contained in $Y$ is more than hand image cropping: it also contains the location of cropping on the image, which can be described by the location of hand detection boxes.

The PIE mechanism aims at infusing this condition into the latent features $\{P_i\}_{i=1}^N$ of image cropping. Given the square cropping box of the hand, we uniformly sample $P\times Q$ points within the cropping as $\{p_{ij}\in\mathbb R^{2}\}^{P\times Q}_{ij}$, where $p_{ij}$ is the pixel coordinate of each sample point. For each sample point, we normalize its pixel coordinate $p_{ij}$ back to $u_{ij}$ using intrinsics:
\begin{equation}
    u_{ij}=K^{-1}[p_{ij},1]^\top\in\mathbb R^3,
\end{equation}
where $K\in\mathbb R^{3\times 3}$ is the intrinsics matrix. The direction vector from the $ij$-th sample on the normalized projection plane to the origin is denoted as $\tilde{u}_{ij}=-u_{ij}/\Vert u_{ij}\Vert_2$. We drop the $z$-coordinates and concatenate all the $\tilde{u}_{ij}$s to get the perspective map
\begin{equation}
    F_\text{persp}=\text{cat}((\tilde{u}_{11})_{xy},\cdots,(\tilde{u}_{PQ})_{xy})\in\mathbb R^{PQ\times 2}.
\end{equation}

As shown in \cref{fig:method-st}(i), the Perspective Information Embedder encodes the perspective map $F_\text{persp}$ into the perspective embedding vector $v_\text{persp}=\text{PIE}(F_\text{persp})\in\mathbb R^d$, where $d$ is the dimension of the backbone image patches. The vector $v_\text{persp}$ is then added to each latent patch $\{P_i\}^N_{i=1}$ to infuse perspective information into local image features. We expect this process to mimic the coupling of wrist position and local pose in the latent space, providing clues for camera-space hand pose in $xy$ plane. For detailed architecture of PIE, please refer to the Appendix.

\subsection{Transformation-Isomorphism Supervision}

Transformation Isomorphism Supervision (TIS) stems from a fundamental geometric insight about perspective projection systems. 
This insight asserts a one-to-one correspondence between scale-rotation transformations in the 2D space and those in the 3D pose space. Specifically, if a transformation is applied to a 2D projection result, a corresponding transformation can be applied to its 3D position such that the new 2D result remains the same as the projection of new 3D position, as shown in \cref{fig:ti-fig}.
A formal specification of these scale-rotation transformations is detailed in the  Appendix.
This phenomenon is attributed to the intrinsic nature of hand geometry and the principles of perspective projection. The interplay between these two factors establishes a consistent relationship of \textbf{depth} variations across 2D and 3D spaces.

Let the pose estimation network $f$ predicts hand pose $\theta,\beta,t$ from image $I$, formulated as $(\theta,\beta,t)=f(I)$. As shown in \cref{fig:ti-fig}, when applying scale-rotation transformation $sr_\text{2d}(\cdot)$ to the image, the resulting pose will change accordingly. This process can be formulated as $sr_\text{pose}(\theta,\beta,t) = f(sr_\text{2d}(I))$, where the subscript denotes the 3D pose space and the 2D space, respectively. We note that $sr(\cdot)$ penetrates through the network: $f\circ sr_\text{2d}=sr_\text{pose}\circ f$. In other words, the scale-rotation transformations in 2D and 3D pose space are in one-to-one correspondence. \cite{10.2312:pg.20251270} denotes such relation as \textit{transformation isomorphism}.
This relation potentially captures the depth information, alleviating the challenge of $z$-coordinate localization.

\begin{figure}[!t]
	\centering
	\includegraphics[width=0.95\linewidth]{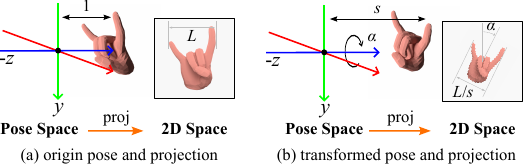}
	\caption{An example of transformation isomorphism. The projection relation holds after applying isomorphic transformations in pose space and image space respectively.}
	\label{fig:ti-fig}
\end{figure}

We naturally conjecture whether this isomorphic relationship between image space and pose space can be extended into the latent space. In our pipeline, the estimation process is formulated as $(\theta,\beta,t)=D(E(I))$. By introducing  transformation $sr_\text{latent}(\cdot)$ to model the ``scale \& rotation'' transformation in the latent space, now the transformation isomorphism relation becomes\footnote{We omitted the $\alpha,s$ and camera intrinsics $K$ for simplicity.}
\begin{equation}
    sr_\text{pose}(\theta,\beta,t)=D(sr_\text{latent}(E(I)))=D(E(sr_\text{2d}(I))).
    \label{eq:trans-iso}
\end{equation}
By enforcing this constraint, TIS encourages the latent space to preserve structurally analogous relationships with both image and pose spaces, facilitating depth information embedding.

\begin{figure}[!t]
    \centering
    \includegraphics[width=0.95\linewidth]{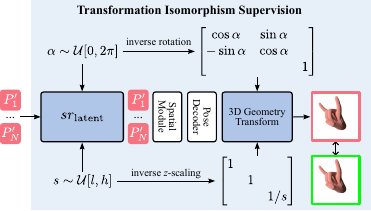}
    \caption{Illustration of Transformation Isomorphism Supervision. A parameterized network $sr_\text{latent}$ takes random rotation and scaling parameters $\alpha, s$ as input to transform the latent patches, and for the pose decoded from the transformed latent patches, it applies inverse 3D rotation and scaling transformations with $\alpha, s$, then compares the results with the truth.}
    \label{fig:tis}
\end{figure}

The mechanism of TIS is shown in \cref{fig:tis}. The parameterized network $sr_{\text{latent}}:\mathbb R^{Nd+2}\to\mathbb R^{Nd}$ is adopted to ``scale'' and ``rotate'' the latent patches $\{P_i'\}_{i=1}^N$ by random $s\sim\mathcal U$ and $\alpha\sim\mathcal U$. The shared decoder $D$ interprets the latent patches into pose $\theta'$, wrist translation $t'$ and shape $\beta'$. By transformation isomorphism relation in \cref{eq:trans-iso}, we expect $sr_\text{pose}(\theta',\beta',t';-\alpha,1/s)=(\theta,\beta,t)$. We construct the objective of TIS as the regularization term with respect to $E,D$ and $sr_\text{latent}$ as:
\begin{equation}
    \mathcal L_\text{TIS}=\mathbb E_{I,sr}\left[ \Vert \text{gt}-(sr_\text{pose})^{-1}(D(sr_\text{latent}(E(I)))) \Vert \right],
\end{equation}
where $(sr_\text{pose})^{-1}(\cdot)$ denotes the inverse of $sr_\text{pose}(\cdot)$, with factor $1/s$ and $-\alpha$. Our ablation studies exhibit the positive effect of TIS.

Unlike \cite{10.2312:pg.20251270}, the transformation isomorphism relations we employ for supervision are established between the latent space and the pose space, whereas \cite{10.2312:pg.20251270} utilizes the relations between the image space and the latent space for supervision.

\subsection{Framerate-Aware Multi-Dataset Training}
Large scale annotated data and temporal information fusion is vital for accurate global hand pose estimation. Incorporating multiple public datasets can solve the first problem, but also introduces the second problem: the capture configuration varies across datasets, especially the framerate. We propose our framerate-aware multi-dataset training (shortened as \textbf{MDT}) approach supporting \textbf{different framerate} in temporal training scenario.

\begin{figure*}[!t]
	\centering
	\includegraphics[width=\linewidth]{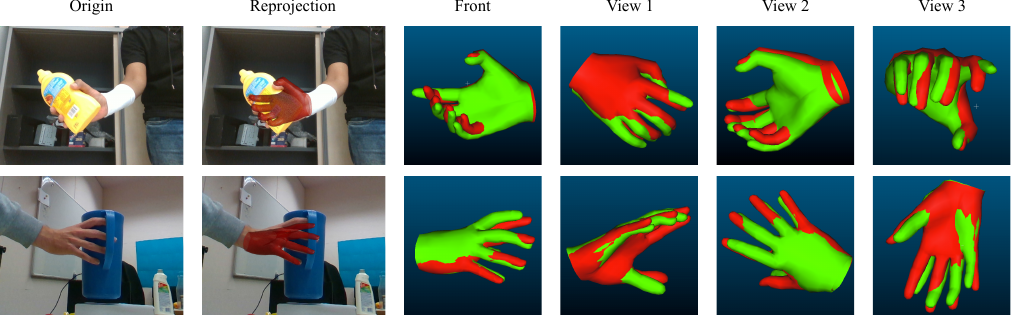}
	\caption{We present qualitative results on the HO3D \cite{Hampali_2020_CVPR} datasets. Ground truth meshes are shown in green, and our predictions are in red. All predictions are rendered in camera space with entirely estimated 3D hand pose (predicted local pose parameter $\theta,\beta$ and wrist position $t$ relative to camera). No ground-truth wrist position or wrist alignment is introduced.}
	\label{fig:qualitative}
\end{figure*}

As shown in \cref{fig:method-st}(ii), given a sequence of images $\{I_i\}^{t}$ with corresponding timestamps $\{\tau_i\}^t$, the spatial module extracts feature sequences for local pose $\{Q_P^i\}^t$, root position $\{Q_T^i\}^t$, and shape $\{Q_S^i\}^t$. We then apply our Time-indexed RoPE (Ti-RoPE, based on RoPE\cite{10.1016/j.neucom.2023.127063}) to these sequences. The last token of each sequence $Q_P^t,Q_T^t,Q_S^t$ is treated as a query to aggregate information from its respective sequence:
\begin{equation}
\begin{aligned}
    \verb|X| &\text{ can be } P,T \text{ and } S, \\
    \{\hat{Q}_{\verb|X|}^{i}\}^t &= \text{Time-Indexed-RoPE}\left(\{Q_{\verb|X|}^i\}^t,\{\tau_i\}^t\right), \\
    \tilde Q_{\verb|X|}^t &= \hat{Q}_{\verb|X|}^t + \text{TM}_{\verb|X|}\left(\verb|q=|\hat{Q}_{\verb|X|}^{t},\verb|v&k=|\{\hat{Q}_{\verb|X|}^{i}\}^t\right),
\end{aligned}
\end{equation}
where $\tilde{Q}_{\verb|X|}^t$ are fused token for pose, wrist translation and shape across time and $\text{TM}_P,\text{TM}_T,\text{TM}_S$ are temporal module branches for local pose, wrist translation and shape feature tokens. Each of them is composed of layers of cross attention module.

Our TI-RoPE is a variant of Rotary Position Embedding \cite{su2021roformer} designed to handle irregularly sampled time-series sequences by incorporating explicit temporal intervals into position encoding. In the original RoPE mechanism, the rotation angle $\theta$ of the embedding is determined solely by the token's index $i$. In our proposed TI-RoPE, we replace this with the \textbf{corresponding timestamp} $\tau_i$ of each token's position, thereby making the encoding sensitive to framerate variations.

For better training efficiency and accuracy, we zero initialize the last linear layer of the Temporal Module, following ControlNet \cite{controlnet:zhang2023adding}. This forces the output of Temporal Module to be zero at the start of the training, aligning the behavior to the pipeline without the Temporal Module.

\subsection{Loss Functions}
The loss function for spatial training is defined as followed:
\begin{equation}
    \mathcal{L}=\mathcal{L}_\text{joint}+\mathcal{L}_\text{vert}+\mathcal{L}_\text{beta}+\mathcal{L}_\text{TIS},
\end{equation}
where $\mathcal{L}_\text{joint}$ and $\mathcal{L}_\text{vert}$ are the L2 loss between the predictions and ground truth for 3D joints and mesh vertices, respectively. $\mathcal{L}_\text{beta}$ is the L1 loss between the predicted shape parameters and the ground truth. The loss function for temporal training is defined as:
\begin{equation}
    \mathcal{L}=\sum_{t=1}^{T}\mathcal{L}_\text{joint}^t+\mathcal{L}_\text{vert}^t+\mathcal{L}_\text{beta}^t,
\end{equation}
where $t$ stands for time index.


\section{Experiments}

\subsection{Implementation Details}
We use Swin-Transformer \cite{liu2021swinv2} with ``base'' scale as our vision backbone. The input image is fixed to $256\times 256$, producing images patches in $\{P_1,\cdots,P_{64}\}\in\mathbb R^{64\times 768}$. For each of $L=6$ spatial modules, the standard self-attention module is implemented, with head number and hidden dimension equal to the configuration of backbone. The detailed architecture about spatial module can be found in \cref{fig:spatial-module}. The perspective information embedding module is a 3-layer MLP. The $sr_\text{latent}$ is composed of 2 standard self-attention layers. For temporal module, three individual branches with 2 cross-attention layers are inserted for pose, shape and translation.

During the spatial training phase, the temporal module are omitted and we update the backbone, spatial module, perspective embedding module and isomorphism transformation $sr_\text{latent}$. We denoted the model trained under this phase as ``Spatial (X)'', where `X' refers the used dataset. In the temporal training phase, the temporal module is inserted and updated solely. The model under this phase is denoted ``Temporal (X)''. The ``MDT'' setup refers the model trained under Temporal phase using all listed datasets. We use AdamW \cite{adamw:loshchilov2019decoupledweightdecayregularization} with learning rate $\sqrt{bs/44}\times 10^{-4}$ and clipping gradients below $5.0$ for training, $bs$ for batch size.
Readers can refer to the Appendix for more training details.

\subsection{Datasets and Metrics}
\paragraph{DexYCB}\cite{dexycb:chao:cvpr2021} is a large-scale RGB‑D dataset capturing diverse natural hand‑object interactions over 582K frames from multi‑view recordings, annotated with accurate 3D hand poses and fingertip keypoints. The timestamps within the sequences need to be consecutive for temporal training, thus we use S1 protocol for our training/test splitting.

\paragraph{HO3D}\cite{Hampali_2020_CVPR} is an RGB‑D dataset focused on realistic hand–object interaction, providing over 103K frames of human manipulation captured under heavy mutual occlusion, each annotated with precise 3D hand poses and corresponding object poses. 

\paragraph{InterHand2.6M}\cite{Moon_2020_ECCV_InterHand2.6M} is a massive-scale RGB dataset for 3D hand pose estimation, distinguished by its extensive coverage of two-hand interactions. It comprises over 2.6 million annotated frames captured from a multi-view camera setup. The dataset provides accurate 3D hand joint annotations for both single-hand and interacting-hand scenarios. 

\paragraph{CS-MJE, RS-MJE, RTE, PA-MJE} Following the metrics mentioned in Huang et al. \cite{nvf:huang_neural_2023}, we report the mean joint error in camera space (CS-MJE), root-aligned joint error (RS-MJE, RS for Root-Space) and wrist translation error in camera space (RTE) to evaluate the performance of our model. Other than that, we report the mean joint error after Procrustes analysis alignment as PA-MJE.

\subsection{Qualitative Result}
Our qualitative results are presented in \cref{fig:qualitative}. We visualize comparisons between our predicted poses (red) and the ground truth (green) from multiple perspectives. Accurate hand pose estimation in camera space, including root joint positioning, is validated by the significant mesh overlap observed in our results.
It is noteworthy that \cref{fig:qualitative} depicts raw network predictions rendered \textbf{with no ground-truth root} or root joint alignment, in contrast to the visualization of local pose works \cite{10655481,deformer:Fu_2023_ICCV,h2onet:Xu_2023_CVPR,potamias2024wilor}. We provide more qualitative results and analyze the influence of root alignment in the supplementary.


\begin{figure}[!t]
    \centering
    \includegraphics[width=\linewidth]{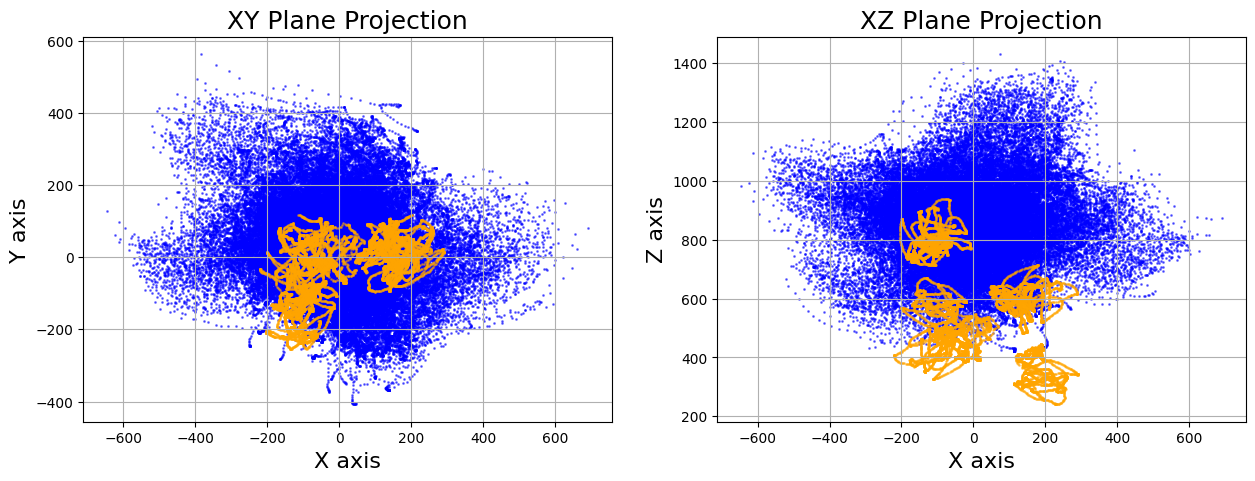}
    \caption{Wrist joint distribution of \textcolor{blue}{DexYCB} and \textcolor{orange}{HO3D}. The poses in the DexYCB dataset significantly surpass those in HO3D in terms of both quantity and range, resulting in greater difficulty for global pose estimation.}
    \label{fig:dataset-dist}
\end{figure}

\begin{table}[!t]
    \centering
    \small
    \caption{Our metrices reported on HO3D. Our model outperforms the SOTA work in the same field \cite{handdgp:leonardis_computer_2025} by 7.8\%, 11.7\% and 37.1\% when training on single frame, multiple frames and multiple datasets respectively. When compared with SOTA work of local hand pose estimation \cite{10655481}, our model still exhibits overall superiority.}
    \label{tab:ho3d}
    \begin{tabular}{lcccc}
        \toprule
         Method & CS-MJE$\downarrow$ & RS-MJE$\downarrow$ & PA-MJE$\downarrow$ & RTE$\downarrow$ \\
         \midrule
         NVF (relative) \cite{nvf:huang_neural_2023} & - & 23.2 & 21.8 & - \\
         HandDGP\cite{handdgp:leonardis_computer_2025} & 50.3 & - & - & - \\
         \midrule
         HaMeR (ViT-H) \cite{10655481} + PnP & \underline{33.8} & \underline{19.1} & \textbf{7.2} & \underline{36.0} \\
         HaMeR (ViT-B) \cite{10655481} + PnP & 55.6 & 35.8 & 11.1 & 48.5 \\
         \midrule
         Spatial (HO3D) & 46.4 & 26.3 & 10.6 & 44.3 \\
         Temporal (HO3D) & 44.4 & 25.9 & 10.6 & 41.9 \\
         Spatial (HO3D+DexYCB) & 43.0 & 23.4 & 9.4 & 39.7 \\
         Temporal (HO3D+DexYCB) & 36.8 & 23.8 & 9.3 & 33.8 \\
         MDT & \textbf{31.7} & \textbf{17.9} & \underline{8.1} & \textbf{31.4} \\
         \bottomrule
    \end{tabular}
\end{table}

\subsection{Quantitative Result}

\paragraph{Quantitative result on HO3D}

\cref{tab:ho3d} reports performance on HO3D \cite{Hampali_2020_CVPR}. Note that NVF was trained with local pose supervision only, whereas camera-space estimation requires additional root joint localization (high-DoF). This methodological discrepancy renders direct comparison unfair; NVF results are thus included solely for reference completeness.

In contrast, HandDGP\cite{handdgp:leonardis_computer_2025} targets the same camera-space task. The proposed method consistently outperforms HandDGP across all configurations by significant margins (e.g., 7.8\% reduction in CS-MJE), demonstrating superior camera-space estimation capability.

To further validate effectiveness against current SOTA of local pose estimation, we compare with HaMeR\cite{10655481}, the work incorporating a large amount of data and huge parameters (ViT-H). Since HaMeR predicts poses in local coordinates, we apply Perspective-$n$-Point (PnP) post-processing to lift predictions to camera space for equitable evaluation. Even though our model and data are not comparable in scale to HaMeR, we still conducted experiments to demonstrate our method's hand geometry modeling capability in camera-space pose estimation. HaMeR is compared with the MDT setup in the following analysis.

As shown in \cref{tab:ho3d}, the proposed method achieves superior CS-MJE, RS-MJE, and RTE, with only marginally higher PA-MJE. This pattern stems from fundamental differences in handling perspective distortion: the proposed method explicitly encodes intrinsics and perspective effects, enabling disentanglement of global position and local pose that are coupled under perspective projection. This yields accurate camera-space localization (lower CS-MJE and RTE) and refined local pose estimation (lower RS-MJE).

The relative disadvantage in PA-MJE arises from a limitation of the metric itself rather than estimation quality. PA-MJE applies Procrustes alignment between predictions and ground truth, which compensates for local pose errors caused by perspective coupling. Consequently, HaMeR benefits from this alignment-based error cancellation, whereas our method—already accounting for perspective effects via explicit intrinsics—gains no such artificial advantage.

To isolate architectural contributions from model capacity effects, we trained a ViT-B variant of HaMeR (matching our backbone scale). With this configuration, all metrics degraded substantially, implying that HaMeR's effectiveness relies more on accurate data distribution fitting than on proper modeling of hand geometry.  

Furthermore, results at varying training data scales demonstrate consistent performance gains with increased data volume, indicating the scalability of the proposed approach.

\begin{table}[!t]
    \centering
    \small
    \caption{Our metrics reported on DexYCB. The evaluation metrics on DexYCB are consistently lower than those on HO3D, primarily due to its broader pose distribution and frequent motion blur in captured images.}
    \label{tab:dexycb}
    \begin{tabular}{lcccc}
        \toprule
         Method & CS-MJE$\downarrow$ & RS-MJE$\downarrow$ & PA-MJE$\downarrow$ & RTE$\downarrow$ \\
         \midrule
         Spatial (DexYCB) & 50.1 & 19.4 & 6.9 & 49.5 \\
         Temporal (DexYCB) & 45.5 & 18.0 & 6.9 & 44.2 \\
         \midrule
         MDT & \textbf{43.7} & \textbf{15.5} & \textbf{6.4} & \textbf{42.7} \\
         \bottomrule
    \end{tabular}
\end{table}

\paragraph{Quantitative result on DexYCB}
To the best of our knowledge, we are the first to report camera-space hand pose estimation performance on DexYCB\cite{dexycb:chao:cvpr2021}. Compared to the HO3D dataset, the DexYCB dataset features a wider range of hand motions, creating a broader spatial distribution (\cref{fig:dataset-dist}) with harder edge cases suffering greater perspective distortion. Our metrics on DexYCB are listed in \cref{tab:dexycb}. Due to wilder pose distribution (shown in \cref{fig:dataset-dist}), the pose estimation task on this dataset is harder. However, no significant accuracy drops are observed for this dataset.

\begin{table}[!t]
    \centering
    \caption{Comparison with our proposed baseline on HO3D. The ``observ.'' denotes the our proposed observation and ``Spatial (HO3D) w/o TIS'' denotes our method without TIS. The table shows the importance of PIE.}
    \label{tab:ablation_baseline}
    \begin{tabular}{lcccc}
        \toprule
         Method & CS-MJE$\downarrow$ & RS-MJE$\downarrow$ & PA-MJE$\downarrow$ & RTE$\downarrow$ \\
        \midrule
        observ. & 49.1 & \textbf{25.1} & 10.5 & 46.4 \\
        Spatial (HO3D) w/o TIS & 46.6 & 25.6 & \textbf{10.4} & 45.1 \\
        Spatial (HO3D) & \textbf{46.4} & 26.3 & 10.6 & \textbf{44.3} \\
        \bottomrule
    \end{tabular}
\end{table}

\subsection{Comparison with Our Observation Baseline}

We proposed a baseline estimating the camera-space hand pose from local hand pose inspired by \cref{fig:task-def-and-challenge} (b,ii). As shown in the picture, different hands with the same distance to the camera and palm orientation towards the camera, produces almost the same projection.
Local hand pose estimation networks, trained on cropped images, implicitly assume a canonical, hand-centric viewpoint. The predicted 3D pose is therefore expressed in a canonical coordinate system. As illustrated in \cref{fig:our-baseline}, we investigate whether applying the corresponding rotation to map this local pose back into the camera space is sufficient to yield an accurate hand pose.

We train the model based on this observation in HO3D and report the metrics, denoted as ``observ.''. Our proposed method is denoted as ``PIE+TIS''. The result can be found on the \cref{tab:ablation_baseline} and \ref{tab:xyz-location}. The proposed baseline deteriorates in camera-space estimation accuracy, proving the importance of perspective information embedding mechanism. The comparison between ``observ.'' and ``PIE'' in \cref{tab:xyz-location} shows that PIE improves the pose estimation in $xy$ plane. The comparison between ``PIE'' and ``TIS+PIE'' proves that TIS is essential for accurate localization on $z$ axis. Despite the negative impact of TIS on $xy$-plane accuracy, improved depth estimation ($z$-axis) compensates for this deficiency, resulting in improvements on the CS-MJE metric, which this task focuses more on (\cref{tab:ablation_baseline}).

\begin{table}[!t]
    \centering
    \caption{Errors in $x$, $y$ and $z$ coordinate for all joints and root joints (with subscript $r$). Evaluated on HO3D.}
    \label{tab:xyz-location}
    \begin{tabular}{l|cc|cc|cc}
        \toprule
         Method & $x$ & $x_r$ & $y$ & $y_r$ & $z$ & $z_r$ \\
         \midrule
         observ. & \textbf{14.87} & 15.37 & 13.57 & 13.26 & 40.28 & 37.78 \\
         PIE & 16.94 & 15.07 & \textbf{13.05} & \textbf{11.48} & 39.97 & 37.01 \\
         PIE+TIS & 15.03 & \textbf{14.81} & 13.91 & 13.65 & \textbf{36.83} & \textbf{35.58} \\
         \bottomrule
    \end{tabular}
\end{table}

\begin{table}[!t]
    \centering
    \caption{Ablation result of TIS mechanism. Adding TIS in the first training stage reduces CS-MJE and RTE.}
    \label{tab:ablation_TIS}
    \begin{tabular}{llccc}
        \toprule
         Method & Dataset & CS-MJE$\downarrow$ & PA-MJE$\downarrow$ & RTE$\downarrow$ \\
         \midrule
         w/o TIS & DexYCB & 51.8 & \textbf{6.9} & 51.5 \\
         w/ TIS & DexYCB & \textbf{50.1} & \textbf{6.9} & \textbf{49.5} \\
         \midrule
         w/o TIS & HO3D & 50.1 & 10.7 & 45.1 \\
         w/ TIS & HO3D & \textbf{46.4} & \textbf{10.6} & \textbf{44.3} \\
         \bottomrule
    \end{tabular}
\end{table}

\subsection{Ablation Studies}

\paragraph{Ablation study about TIS}
TIS enhances the model by improving the depth sensitivity of backbone. We conducted ablation study on TIS mechanism. As shown in \cref{tab:ablation_TIS}, after removing TIS mechanism from the spatial training, the position error of global hand pose estimation increases by $3.2\%$ and $7.4\%$ on DexYCB and HO3D. It proves that by introducing latent transformation that corresponding to the ones in the 2D and 3D space, we can boost the backbone for the extraction of depth information. Although the TIS mechanism improves the global positioning ability, it does not negatively affect the local hand pose estimation, keeping the PA-MJE performance almost unchanged.

\paragraph{Ablation study about PIE mechanism}
We ablate the implementation of PIE mechanism. In our proposed PIE mechanism, we combine the perspective embedding vector $v_\text{persp}$ with the sequence of image patches $\{P_i\}_i^N$ through summation, denoted as ``addpatch''. For ablation, we added $v_\text{persp}$ to pose query token $Q_P^+,Q_T^+,Q_S^+$, denoted as ``addquery''. We also ablate the implementation of PIE input construction. Originally, we utilize dense perspective direction map $F_\text{persp}$ to represent the perspective information. We explore using the normalized coordinates of four corner points for the embedding of the perspective information, denoted as ``sparse''.

The metrics of these ablations on HO3D are shown in \cref{tab:ablation_PIE}. The table shows that ``addpatch'' exhibits best performance, demonstrating that integrating dense perspective information with image features improves camera-space localization accuracy.

\begin{table}[!t]
    \centering
    \caption{Ablation result of PIE mechanism. We evaluate the metrics on HO3D.}
    \label{tab:ablation_PIE}
    \begin{tabular}{lcccc}
        \toprule
         Method & CS-MJE$\downarrow$ & RS-MJE$\downarrow$ & PA-MJE$\downarrow$ & RTE$\downarrow$ \\
         \midrule
         addquery & 48.7 & 27.2 & 10.7 & 47.1 \\
         sparse & 46.5 & 26.8 & 10.7 & 44.4 \\
         addpatch & \textbf{46.4} & \textbf{26.3} & \textbf{10.6} & \textbf{44.3} \\
         \bottomrule
    \end{tabular}
\end{table}

\begin{table}[!t]
    \centering
    \caption{Ablation study of zero initialization mechanism in framerate-aware multi-dataset training. The comparison exhibits positive effect of our proposed method.}
    \label{tab:ablation_zero}
    \begin{tabular}{lcccc}
        \toprule
         Method & CS-MJE$\downarrow$ & RS-MJE$\downarrow$ & PA-MJE$\downarrow$ & RTE$\downarrow$ \\
        \midrule
        random & 47.9 & 26.4 & 10.7 & 43.8 \\
        zero & \textbf{44.8} & \textbf{25.8} & \textbf{10.6} & \textbf{42.4} \\
        \bottomrule
    \end{tabular}
\end{table}

\paragraph{Ablation study about zero initialization}

The zero-initialization of last linear layer in temporal module makes entire pipeline behave as if there is no temporal module at the beginning of temporal training, as the output of temporal module is zeroed-out and hence has no influence on pose, shape and translation token from the spatial module. This improves the performance compared to the random initialization, as shown in \cref{tab:ablation_zero}. It proves that providing a fluent transition between stages is crucial for both training and inference.


\section{Conclusion}

We proposed a novel approach for accurate hand pose estimation in camera space. This method can enhance the quality of existing interaction devices by providing pose with respective to camera. The proposed TIS, PIE and MDT mechanism have proved their effectiveness through ablation studies and outperformed the SOTA both in local and camera-space pose estimation task. Further works on effectively utilizing distortion information and temporal clues are under investigation.


\begin{acks}
This work was supported by The Innovation Funding of Institute of Computing Technology, Chinese Academy of Sciences (Grant No. E561010) and the National Key Research and Development Program of China (Grant No. 2022YFB3303202).
\end{acks}

\bibliographystyle{ACM-Reference-Format}
\balance
\bibliography{main}


\begin{thebibliography}{40}


\ifx \showCODEN    \undefined \def \showCODEN     #1{\unskip}     \fi
\ifx \showISBNx    \undefined \def \showISBNx     #1{\unskip}     \fi
\ifx \showISBNxiii \undefined \def \showISBNxiii  #1{\unskip}     \fi
\ifx \showISSN     \undefined \def \showISSN      #1{\unskip}     \fi
\ifx \showLCCN     \undefined \def \showLCCN      #1{\unskip}     \fi
\ifx \shownote     \undefined \def \shownote      #1{#1}          \fi
\ifx \showarticletitle \undefined \def \showarticletitle #1{#1}   \fi
\ifx \showURL      \undefined \def \showURL       {\relax}        \fi
\providecommand\bibfield[2]{#2}
\providecommand\bibinfo[2]{#2}
\providecommand\natexlab[1]{#1}
\providecommand\showeprint[2][]{arXiv:#2}

\bibitem[Chao et~al\mbox{.}(2021)]%
        {dexycb:chao:cvpr2021}
\bibfield{author}{\bibinfo{person}{Yu-Wei Chao}, \bibinfo{person}{Wei Yang}, \bibinfo{person}{Yu Xiang}, \bibinfo{person}{Pavlo Molchanov}, \bibinfo{person}{Ankur Handa}, \bibinfo{person}{Jonathan Tremblay}, \bibinfo{person}{Yashraj~S. Narang}, \bibinfo{person}{Karl {Van Wyk}}, \bibinfo{person}{Umar Iqbal}, \bibinfo{person}{Stan Birchfield}, \bibinfo{person}{Jan Kautz}, {and} \bibinfo{person}{Dieter Fox}.} \bibinfo{year}{2021}\natexlab{}.
\newblock \showarticletitle{{DexYCB}: A Benchmark for Capturing Hand Grasping of Objects}. In \bibinfo{booktitle}{\emph{IEEE/CVF Conference on Computer Vision and Pattern Recognition (CVPR)}}.
\newblock


\bibitem[Chen et~al\mbox{.}(2020a)]%
        {pmlr-v119-chen20j}
\bibfield{author}{\bibinfo{person}{Ting Chen}, \bibinfo{person}{Simon Kornblith}, \bibinfo{person}{Mohammad Norouzi}, {and} \bibinfo{person}{Geoffrey Hinton}.} \bibinfo{year}{2020}\natexlab{a}.
\newblock \showarticletitle{A Simple Framework for Contrastive Learning of Visual Representations}. In \bibinfo{booktitle}{\emph{Proceedings of the 37th International Conference on Machine Learning}} \emph{(\bibinfo{series}{Proceedings of Machine Learning Research}, Vol.~\bibinfo{volume}{119})}, \bibfield{editor}{\bibinfo{person}{Hal~Daumé III} {and} \bibinfo{person}{Aarti Singh}} (Eds.). \bibinfo{publisher}{PMLR}, \bibinfo{pages}{1597--1607}.
\newblock
\urldef\tempurl%
\url{https://proceedings.mlr.press/v119/chen20j.html}
\showURL{%
\tempurl}


\bibitem[Chen et~al\mbox{.}(2020b)]%
        {chen2020simple}
\bibfield{author}{\bibinfo{person}{Ting Chen}, \bibinfo{person}{Simon Kornblith}, \bibinfo{person}{Mohammad Norouzi}, {and} \bibinfo{person}{Geoffrey Hinton}.} \bibinfo{year}{2020}\natexlab{b}.
\newblock \showarticletitle{A Simple Framework for Contrastive Learning of Visual Representations}.
\newblock \bibinfo{journal}{\emph{arXiv preprint arXiv:2002.05709}} (\bibinfo{year}{2020}).
\newblock


\bibitem[Contributors(2020)]%
        {mmpose2020}
\bibfield{author}{\bibinfo{person}{MMPose Contributors}.} \bibinfo{year}{2020}\natexlab{}.
\newblock \bibinfo{title}{OpenMMLab Pose Estimation Toolbox and Benchmark}.
\newblock \bibinfo{howpublished}{\url{https://github.com/open-mmlab/mmpose}}.
\newblock


\bibitem[Fu et~al\mbox{.}(2023)]%
        {deformer:Fu_2023_ICCV}
\bibfield{author}{\bibinfo{person}{Qichen Fu}, \bibinfo{person}{Xingyu Liu}, \bibinfo{person}{Ran Xu}, \bibinfo{person}{Juan~Carlos Niebles}, {and} \bibinfo{person}{Kris~M. Kitani}.} \bibinfo{year}{2023}\natexlab{}.
\newblock \showarticletitle{Deformer: Dynamic Fusion Transformer for Robust Hand Pose Estimation}. In \bibinfo{booktitle}{\emph{Proceedings of the IEEE/CVF International Conference on Computer Vision (ICCV)}}. \bibinfo{pages}{23600--23611}.
\newblock


\bibitem[Ge et~al\mbox{.}(2019)]%
        {ge2019handshapepose}
\bibfield{author}{\bibinfo{person}{Liuhao Ge}, \bibinfo{person}{Zhou Ren}, \bibinfo{person}{Yuncheng Li}, \bibinfo{person}{Zehao Xue}, \bibinfo{person}{Yingying Wang}, \bibinfo{person}{Jianfei Cai}, {and} \bibinfo{person}{Junsong Yuan}.} \bibinfo{year}{2019}\natexlab{}.
\newblock \showarticletitle{3D Hand Shape and Pose Estimation from a Single RGB Image}. In \bibinfo{booktitle}{\emph{CVPR}}.
\newblock


\bibitem[Hampali et~al\mbox{.}(2020)]%
        {Hampali_2020_CVPR}
\bibfield{author}{\bibinfo{person}{Shreyas Hampali}, \bibinfo{person}{Mahdi Rad}, \bibinfo{person}{Markus Oberweger}, {and} \bibinfo{person}{Vincent Lepetit}.} \bibinfo{year}{2020}\natexlab{}.
\newblock \showarticletitle{HOnnotate: A Method for 3D Annotation of Hand and Object Poses}. In \bibinfo{booktitle}{\emph{Proceedings of the IEEE/CVF Conference on Computer Vision and Pattern Recognition (CVPR)}}.
\newblock


\bibitem[Hampali et~al\mbox{.}(2022)]%
        {Hampali_2022_CVPR_Kypt_Trans}
\bibfield{author}{\bibinfo{person}{Shreyas Hampali}, \bibinfo{person}{Sayan~Deb Sarkar}, \bibinfo{person}{Mahdi Rad}, {and} \bibinfo{person}{Vincent Lepetit}.} \bibinfo{year}{2022}\natexlab{}.
\newblock \showarticletitle{Keypoint Transformer: Solving Joint Identification in Challenging Hands and Object Interactions for Accurate 3D Pose Estimation}. In \bibinfo{booktitle}{\emph{IEEE Computer Vision and Pattern Recognition Conference}}.
\newblock


\bibitem[He et~al\mbox{.}(2021)]%
        {MaskedAutoencoders2021}
\bibfield{author}{\bibinfo{person}{Kaiming He}, \bibinfo{person}{Xinlei Chen}, \bibinfo{person}{Saining Xie}, \bibinfo{person}{Yanghao Li}, \bibinfo{person}{Piotr Doll{\'a}r}, {and} \bibinfo{person}{Ross Girshick}.} \bibinfo{year}{2021}\natexlab{}.
\newblock \showarticletitle{Masked Autoencoders Are Scalable Vision Learners}.
\newblock \bibinfo{journal}{\emph{arXiv:2111.06377}} (\bibinfo{year}{2021}).
\newblock


\bibitem[Huang et~al\mbox{.}(2023)]%
        {nvf:huang_neural_2023}
\bibfield{author}{\bibinfo{person}{Lin Huang}, \bibinfo{person}{Chung-Ching Lin}, \bibinfo{person}{Kevin Lin}, \bibinfo{person}{Lin Liang}, \bibinfo{person}{Lijuan Wang}, \bibinfo{person}{Junsong Yuan}, {and} \bibinfo{person}{Zicheng Liu}.} \bibinfo{year}{2023}\natexlab{}.
\newblock \showarticletitle{Neural {Voting} {Field} for {Camera}-{Space} {3D} {Hand} {Pose} {Estimation}}. In \bibinfo{booktitle}{\emph{2023 {IEEE}/{CVF} {Conference} on {Computer} {Vision} and {Pattern} {Recognition} ({CVPR})}}. \bibinfo{publisher}{IEEE}, \bibinfo{address}{Vancouver, BC, Canada}, \bibinfo{pages}{8969--8978}.
\newblock
\showISBNx{9798350301298}
\href{https://doi.org/10.1109/CVPR52729.2023.00866}{doi:\nolinkurl{10.1109/CVPR52729.2023.00866}}


\bibitem[Kaplan et~al\mbox{.}(2020)]%
        {kaplan2020scalinglawsneurallanguage}
\bibfield{author}{\bibinfo{person}{Jared Kaplan}, \bibinfo{person}{Sam McCandlish}, \bibinfo{person}{Tom Henighan}, \bibinfo{person}{Tom~B. Brown}, \bibinfo{person}{Benjamin Chess}, \bibinfo{person}{Rewon Child}, \bibinfo{person}{Scott Gray}, \bibinfo{person}{Alec Radford}, \bibinfo{person}{Jeffrey Wu}, {and} \bibinfo{person}{Dario Amodei}.} \bibinfo{year}{2020}\natexlab{}.
\newblock \bibinfo{title}{Scaling Laws for Neural Language Models}.
\newblock
\showeprint[arxiv]{2001.08361}~[cs.LG]
\urldef\tempurl%
\url{https://arxiv.org/abs/2001.08361}
\showURL{%
\tempurl}


\bibitem[Leonardis et~al\mbox{.}(2025)]%
        {handdgp:leonardis_computer_2025}
\bibfield{editor}{\bibinfo{person}{Aleš Leonardis}, \bibinfo{person}{Elisa Ricci}, \bibinfo{person}{Stefan Roth}, \bibinfo{person}{Olga Russakovsky}, \bibinfo{person}{Torsten Sattler}, {and} \bibinfo{person}{Gül Varol}} (Eds.). \bibinfo{year}{2025}\natexlab{}.
\newblock \bibinfo{booktitle}{\emph{Computer {Vision} – {ECCV} 2024: 18th {European} {Conference}, {Milan}, {Italy}, {September} 29–{October} 4, 2024, {Proceedings}, {Part} {XXXVIII}}}. \bibinfo{series}{Lecture {Notes} in {Computer} {Science}}, Vol.~\bibinfo{volume}{15096}.
\newblock \bibinfo{publisher}{Springer Nature Switzerland}, \bibinfo{address}{Cham}.
\newblock
\showISBNx{978-3-031-72919-5 978-3-031-72920-1}
\href{https://doi.org/10.1007/978-3-031-72920-1}{doi:\nolinkurl{10.1007/978-3-031-72920-1}}


\bibitem[Lin et~al\mbox{.}(2025)]%
        {lin2025simhand}
\bibfield{author}{\bibinfo{person}{Nie Lin}, \bibinfo{person}{Takehiko Ohkawa}, \bibinfo{person}{Yifei Huang}, \bibinfo{person}{Mingfang Zhang}, \bibinfo{person}{Minjie Cai}, \bibinfo{person}{Ming Li}, \bibinfo{person}{Ryosuke Furuta}, {and} \bibinfo{person}{Yoichi Sato}.} \bibinfo{year}{2025}\natexlab{}.
\newblock \showarticletitle{Si{MH}and: Mining Similar Hands for Large-Scale 3D Hand Pose Pre-training}. In \bibinfo{booktitle}{\emph{The Thirteenth International Conference on Learning Representations}}.
\newblock
\urldef\tempurl%
\url{https://openreview.net/forum?id=96jZFqM5E0}
\showURL{%
\tempurl}


\bibitem[Lin et~al\mbox{.}(2024)]%
        {lin_pre-training_2024}
\bibfield{author}{\bibinfo{person}{Nie Lin}, \bibinfo{person}{Takehiko Ohkawa}, \bibinfo{person}{Mingfang Zhang}, \bibinfo{person}{Yifei Huang}, \bibinfo{person}{Ryosuke Furuta}, {and} \bibinfo{person}{Yoichi Sato}.} \bibinfo{year}{2024}\natexlab{}.
\newblock \bibinfo{title}{Pre-{Training} for {3D} {Hand} {Pose} {Estimation} with {Contrastive} {Learning} on {Large}-{Scale} {Hand} {Images} in the {Wild}}.
\newblock
\shownote{arXiv:2409.09714 [cs]}.
\newblock
\urldef\tempurl%
\url{http://arxiv.org/abs/2409.09714}
\showURL{%
\tempurl}


\bibitem[Liu et~al\mbox{.}(2022)]%
        {liu2021swinv2}
\bibfield{author}{\bibinfo{person}{Ze Liu}, \bibinfo{person}{Han Hu}, \bibinfo{person}{Yutong Lin}, \bibinfo{person}{Zhuliang Yao}, \bibinfo{person}{Zhenda Xie}, \bibinfo{person}{Yixuan Wei}, \bibinfo{person}{Jia Ning}, \bibinfo{person}{Yue Cao}, \bibinfo{person}{Zheng Zhang}, \bibinfo{person}{Li Dong}, \bibinfo{person}{Furu Wei}, {and} \bibinfo{person}{Baining Guo}.} \bibinfo{year}{2022}\natexlab{}.
\newblock \showarticletitle{Swin Transformer V2: Scaling Up Capacity and Resolution}. In \bibinfo{booktitle}{\emph{International Conference on Computer Vision and Pattern Recognition (CVPR)}}.
\newblock


\bibitem[Loshchilov and Hutter(2019)]%
        {adamw:loshchilov2019decoupledweightdecayregularization}
\bibfield{author}{\bibinfo{person}{Ilya Loshchilov} {and} \bibinfo{person}{Frank Hutter}.} \bibinfo{year}{2019}\natexlab{}.
\newblock \bibinfo{title}{Decoupled Weight Decay Regularization}.
\newblock
\showeprint[arxiv]{1711.05101}~[cs.LG]
\urldef\tempurl%
\url{https://arxiv.org/abs/1711.05101}
\showURL{%
\tempurl}


\bibitem[Lugaresi et~al\mbox{.}(2019)]%
        {Lugaresi2019MediaPipeAF}
\bibfield{author}{\bibinfo{person}{Camillo Lugaresi}, \bibinfo{person}{Jiuqiang Tang}, \bibinfo{person}{Hadon Nash}, \bibinfo{person}{Chris McClanahan}, \bibinfo{person}{Esha Uboweja}, \bibinfo{person}{Michael Hays}, \bibinfo{person}{Fan Zhang}, \bibinfo{person}{Chuo-Ling Chang}, \bibinfo{person}{Ming~Guang Yong}, \bibinfo{person}{Juhyun Lee}, \bibinfo{person}{Wan-Teh Chang}, \bibinfo{person}{Wei Hua}, \bibinfo{person}{Manfred Georg}, {and} \bibinfo{person}{Matthias Grundmann}.} \bibinfo{year}{2019}\natexlab{}.
\newblock \showarticletitle{MediaPipe: A Framework for Building Perception Pipelines}.
\newblock \bibinfo{journal}{\emph{ArXiv}}  \bibinfo{volume}{abs/1906.08172} (\bibinfo{year}{2019}).
\newblock
\urldef\tempurl%
\url{https://api.semanticscholar.org/CorpusID:195069430}
\showURL{%
\tempurl}


\bibitem[Meng et~al\mbox{.}(2022)]%
        {10.1007/978-3-031-20068-7_22}
\bibfield{author}{\bibinfo{person}{Hao Meng}, \bibinfo{person}{Sheng Jin}, \bibinfo{person}{Wentao Liu}, \bibinfo{person}{Chen Qian}, \bibinfo{person}{Mengxiang Lin}, \bibinfo{person}{Wanli Ouyang}, {and} \bibinfo{person}{Ping Luo}.} \bibinfo{year}{2022}\natexlab{}.
\newblock \showarticletitle{3D Interacting Hand Pose Estimation by Hand De-occlusion and Removal}. In \bibinfo{booktitle}{\emph{Computer Vision – ECCV 2022: 17th European Conference, Tel Aviv, Israel, October 23–27, 2022, Proceedings, Part VI}} (Tel Aviv, Israel). \bibinfo{publisher}{Springer-Verlag}, \bibinfo{address}{Berlin, Heidelberg}, \bibinfo{pages}{380–397}.
\newblock
\showISBNx{978-3-031-20067-0}
\href{https://doi.org/10.1007/978-3-031-20068-7_22}{doi:\nolinkurl{10.1007/978-3-031-20068-7_22}}


\bibitem[Moon(2023)]%
        {moon_bringing_2023}
\bibfield{author}{\bibinfo{person}{Gyeongsik Moon}.} \bibinfo{year}{2023}\natexlab{}.
\newblock \bibinfo{title}{Bringing {Inputs} to {Shared} {Domains} for {3D} {Interacting} {Hands} {Recovery} in the {Wild}}.
\newblock
\shownote{arXiv:2303.13652 [cs]}.
\newblock
\urldef\tempurl%
\url{http://arxiv.org/abs/2303.13652}
\showURL{%
\tempurl}


\bibitem[Moon et~al\mbox{.}(2020)]%
        {Moon_2020_ECCV_InterHand2.6M}
\bibfield{author}{\bibinfo{person}{Gyeongsik Moon}, \bibinfo{person}{Shoou-I Yu}, \bibinfo{person}{He Wen}, \bibinfo{person}{Takaaki Shiratori}, {and} \bibinfo{person}{Kyoung~Mu Lee}.} \bibinfo{year}{2020}\natexlab{}.
\newblock \showarticletitle{InterHand2.6M: A Dataset and Baseline for 3D Interacting Hand Pose Estimation from a Single RGB Image}. In \bibinfo{booktitle}{\emph{European Conference on Computer Vision (ECCV)}}.
\newblock


\bibitem[Oh et~al\mbox{.}(2023)]%
        {blur_oh2023recovering3dhandmesh}
\bibfield{author}{\bibinfo{person}{Yeonguk Oh}, \bibinfo{person}{JoonKyu Park}, \bibinfo{person}{Jaeha Kim}, \bibinfo{person}{Gyeongsik Moon}, {and} \bibinfo{person}{Kyoung~Mu Lee}.} \bibinfo{year}{2023}\natexlab{}.
\newblock \bibinfo{title}{Recovering 3D Hand Mesh Sequence from a Single Blurry Image: A New Dataset and Temporal Unfolding}.
\newblock
\showeprint[arxiv]{2303.15417}~[cs.CV]
\urldef\tempurl%
\url{https://arxiv.org/abs/2303.15417}
\showURL{%
\tempurl}


\bibitem[Pavlakos et~al\mbox{.}(2024)]%
        {10655481}
\bibfield{author}{\bibinfo{person}{Georgios Pavlakos}, \bibinfo{person}{Dandan Shan}, \bibinfo{person}{Ilija Radosavovic}, \bibinfo{person}{Angjoo Kanazawa}, \bibinfo{person}{David Fouhey}, {and} \bibinfo{person}{Jitendra Malik}.} \bibinfo{year}{2024}\natexlab{}.
\newblock \showarticletitle{Reconstructing Hands in 3D with Transformers}. In \bibinfo{booktitle}{\emph{2024 IEEE/CVF Conference on Computer Vision and Pattern Recognition (CVPR)}}. \bibinfo{pages}{9826--9836}.
\newblock
\href{https://doi.org/10.1109/CVPR52733.2024.00938}{doi:\nolinkurl{10.1109/CVPR52733.2024.00938}}


\bibitem[Potamias et~al\mbox{.}(2024)]%
        {potamias2024wilor}
\bibfield{author}{\bibinfo{person}{Rolandos~Alexandros Potamias}, \bibinfo{person}{Jinglei Zhang}, \bibinfo{person}{Jiankang Deng}, {and} \bibinfo{person}{Stefanos Zafeiriou}.} \bibinfo{year}{2024}\natexlab{}.
\newblock \bibinfo{title}{WiLoR: End-to-end 3D Hand Localization and Reconstruction in-the-wild}.
\newblock
\showeprint[arxiv]{2409.12259}~[cs.CV]


\bibitem[Potamias et~al\mbox{.}(2025)]%
        {Potamias_2025_CVPR}
\bibfield{author}{\bibinfo{person}{Rolandos~Alexandros Potamias}, \bibinfo{person}{Jinglei Zhang}, \bibinfo{person}{Jiankang Deng}, {and} \bibinfo{person}{Stefanos Zafeiriou}.} \bibinfo{year}{2025}\natexlab{}.
\newblock \showarticletitle{WiLoR: End-to-end 3D Hand Localization and Reconstruction in-the-wild}. In \bibinfo{booktitle}{\emph{Proceedings of the Computer Vision and Pattern Recognition Conference (CVPR)}}. \bibinfo{pages}{12242--12254}.
\newblock


\bibitem[Prakash et~al\mbox{.}(2024)]%
        {ego_Prakash2024Hands}
\bibfield{author}{\bibinfo{person}{Aditya Prakash}, \bibinfo{person}{Ruisen Tu}, \bibinfo{person}{Matthew Chang}, {and} \bibinfo{person}{Saurabh Gupta}.} \bibinfo{year}{2024}\natexlab{}.
\newblock \showarticletitle{3D Hand Pose Estimation in Everyday Egocentric Images}. In \bibinfo{booktitle}{\emph{European Conference on Computer Vision (ECCV)}}.
\newblock


\bibitem[Ren et~al\mbox{.}(2025)]%
        {10.2312:pg.20251270}
\bibfield{author}{\bibinfo{person}{Kaiwen Ren}, \bibinfo{person}{Lei Hu}, \bibinfo{person}{Zhiheng Zhang}, \bibinfo{person}{Yongjing Ye}, {and} \bibinfo{person}{Shihong Xia}.} \bibinfo{year}{2025}\natexlab{}.
\newblock \showarticletitle{{Learning Transformation-Isomorphic Latent Space for Accurate Hand Pose Estimation}}. In \bibinfo{booktitle}{\emph{Pacific Graphics Conference Papers, Posters, and Demos}}, \bibfield{editor}{\bibinfo{person}{Marc Christie}, \bibinfo{person}{Ping-Hsuan Han}, \bibinfo{person}{Shih-Syun Lin}, \bibinfo{person}{Nico Pietroni}, \bibinfo{person}{Teseo Schneider}, \bibinfo{person}{Hsin-Ruey Tsai}, \bibinfo{person}{Yu-Shuen Wang}, {and} \bibinfo{person}{Eugene Zhang}} (Eds.). \bibinfo{publisher}{The Eurographics Association}.
\newblock
\showISBNx{978-3-03868-295-0}
\href{https://doi.org/10.2312/pg.20251270}{doi:\nolinkurl{10.2312/pg.20251270}}


\bibitem[Romero et~al\mbox{.}(2017)]%
        {MANO:SIGGRAPHASIA:2017}
\bibfield{author}{\bibinfo{person}{Javier Romero}, \bibinfo{person}{Dimitrios Tzionas}, {and} \bibinfo{person}{Michael~J. Black}.} \bibinfo{year}{2017}\natexlab{}.
\newblock \showarticletitle{Embodied Hands: Modeling and Capturing Hands and Bodies Together}.
\newblock \bibinfo{journal}{\emph{ACM Transactions on Graphics, (Proc. SIGGRAPH Asia)}} \bibinfo{volume}{36}, \bibinfo{number}{6} (\bibinfo{date}{Nov.} \bibinfo{year}{2017}).
\newblock


\bibitem[Singh et~al\mbox{.}(2024)]%
        {singh_explainable_2024}
\bibfield{author}{\bibinfo{person}{Devansh Singh}, \bibinfo{person}{Aboli Marathe}, \bibinfo{person}{Sidharth Roy}, \bibinfo{person}{Rahee Walambe}, {and} \bibinfo{person}{Ketan Kotecha}.} \bibinfo{year}{2024}\natexlab{}.
\newblock \showarticletitle{Explainable rotation-invariant self-supervised representation learning}.
\newblock \bibinfo{journal}{\emph{MethodsX}}  \bibinfo{volume}{13} (\bibinfo{date}{Dec.} \bibinfo{year}{2024}), \bibinfo{pages}{102959}.
\newblock
\showISSN{22150161}
\href{https://doi.org/10.1016/j.mex.2024.102959}{doi:\nolinkurl{10.1016/j.mex.2024.102959}}


\bibitem[Spurr et~al\mbox{.}(2022)]%
        {spurr_peclr_2022}
\bibfield{author}{\bibinfo{person}{Adrian Spurr}, \bibinfo{person}{Aneesh Dahiya}, \bibinfo{person}{Xi Wang}, \bibinfo{person}{Xucong Zhang}, {and} \bibinfo{person}{Otmar Hilliges}.} \bibinfo{year}{2022}\natexlab{}.
\newblock \bibinfo{title}{{PeCLR}: {Self}-{Supervised} {3D} {Hand} {Pose} {Estimation} from monocular {RGB} via {Equivariant} {Contrastive} {Learning}}.
\newblock
\shownote{arXiv:2106.05953 [cs]}.
\newblock
\urldef\tempurl%
\url{http://arxiv.org/abs/2106.05953}
\showURL{%
\tempurl}


\bibitem[Su et~al\mbox{.}(2024)]%
        {10.1016/j.neucom.2023.127063}
\bibfield{author}{\bibinfo{person}{Jianlin Su}, \bibinfo{person}{Murtadha Ahmed}, \bibinfo{person}{Yu Lu}, \bibinfo{person}{Shengfeng Pan}, \bibinfo{person}{Wen Bo}, {and} \bibinfo{person}{Yunfeng Liu}.} \bibinfo{year}{2024}\natexlab{}.
\newblock \showarticletitle{RoFormer: Enhanced transformer with Rotary Position Embedding}.
\newblock \bibinfo{journal}{\emph{Neurocomput.}} \bibinfo{volume}{568}, \bibinfo{number}{C} (\bibinfo{date}{Feb.} \bibinfo{year}{2024}), \bibinfo{numpages}{12}~pages.
\newblock
\showISSN{0925-2312}
\href{https://doi.org/10.1016/j.neucom.2023.127063}{doi:\nolinkurl{10.1016/j.neucom.2023.127063}}


\bibitem[Su et~al\mbox{.}(2021)]%
        {su2021roformer}
\bibfield{author}{\bibinfo{person}{Jianlin Su}, \bibinfo{person}{Yu Lu}, \bibinfo{person}{Shengfeng Pan}, \bibinfo{person}{Bo Wen}, {and} \bibinfo{person}{Yunfeng Liu}.} \bibinfo{year}{2021}\natexlab{}.
\newblock \bibinfo{title}{RoFormer: Enhanced Transformer with Rotary Position Embedding}.
\newblock
\showeprint[arxiv]{2104.09864}~[cs.CL]


\bibitem[Tu et~al\mbox{.}(2023)]%
        {10.1109/TPAMI.2023.3247907}
\bibfield{author}{\bibinfo{person}{Zhigang Tu}, \bibinfo{person}{Zhisheng Huang}, \bibinfo{person}{Yujin Chen}, \bibinfo{person}{Di Kang}, \bibinfo{person}{Linchao Bao}, \bibinfo{person}{Bisheng Yang}, {and} \bibinfo{person}{Junsong Yuan}.} \bibinfo{year}{2023}\natexlab{}.
\newblock \showarticletitle{Consistent 3D Hand Reconstruction in Video via Self-Supervised Learning}.
\newblock \bibinfo{journal}{\emph{IEEE Trans. Pattern Anal. Mach. Intell.}} \bibinfo{volume}{45}, \bibinfo{number}{8} (\bibinfo{date}{Aug.} \bibinfo{year}{2023}), \bibinfo{pages}{9469–9485}.
\newblock
\showISSN{0162-8828}
\href{https://doi.org/10.1109/TPAMI.2023.3247907}{doi:\nolinkurl{10.1109/TPAMI.2023.3247907}}


\bibitem[Xu et~al\mbox{.}(2023)]%
        {h2onet:Xu_2023_CVPR}
\bibfield{author}{\bibinfo{person}{Hao Xu}, \bibinfo{person}{Tianyu Wang}, \bibinfo{person}{Xiao Tang}, {and} \bibinfo{person}{Chi-Wing Fu}.} \bibinfo{year}{2023}\natexlab{}.
\newblock \showarticletitle{H2ONet: Hand-Occlusion-and-Orientation-Aware Network for Real-Time 3D Hand Mesh Reconstruction}. In \bibinfo{booktitle}{\emph{Proceedings of the IEEE/CVF Conference on Computer Vision and Pattern Recognition (CVPR)}}. \bibinfo{pages}{17048--17058}.
\newblock


\bibitem[Xu et~al\mbox{.}(2022a)]%
        {xu2022vitpose}
\bibfield{author}{\bibinfo{person}{Yufei Xu}, \bibinfo{person}{Jing Zhang}, \bibinfo{person}{Qiming Zhang}, {and} \bibinfo{person}{Dacheng Tao}.} \bibinfo{year}{2022}\natexlab{a}.
\newblock \showarticletitle{Vi{TP}ose: Simple Vision Transformer Baselines for Human Pose Estimation}. In \bibinfo{booktitle}{\emph{Advances in Neural Information Processing Systems}}.
\newblock


\bibitem[Xu et~al\mbox{.}(2022b)]%
        {xu2022vitpose+}
\bibfield{author}{\bibinfo{person}{Yufei Xu}, \bibinfo{person}{Jing Zhang}, \bibinfo{person}{Qiming Zhang}, {and} \bibinfo{person}{Dacheng Tao}.} \bibinfo{year}{2022}\natexlab{b}.
\newblock \showarticletitle{ViTPose+: Vision Transformer Foundation Model for Generic Body Pose Estimation}.
\newblock \bibinfo{journal}{\emph{arXiv preprint arXiv:2212.04246}} (\bibinfo{year}{2022}).
\newblock


\bibitem[Yu et~al\mbox{.}(2024)]%
        {yu2024dynhamr}
\bibfield{author}{\bibinfo{person}{Zhengdi Yu}, \bibinfo{person}{Stefanos Zafeiriou}, {and} \bibinfo{person}{Tolga Birdal}.} \bibinfo{year}{2024}\natexlab{}.
\newblock \showarticletitle{Dyn-HaMR: Recovering 4D Interacting Hand Motion from a Dynamic Camera}. In \bibinfo{booktitle}{\emph{arXiv preprint arXiv:2412.12861}}.
\newblock


\bibitem[Zhang et~al\mbox{.}(2025)]%
        {zhang2025hawor}
\bibfield{author}{\bibinfo{person}{Jinglei Zhang}, \bibinfo{person}{Jiankang Deng}, \bibinfo{person}{Chao Ma}, {and} \bibinfo{person}{Rolandos~Alexandros Potamias}.} \bibinfo{year}{2025}\natexlab{}.
\newblock \showarticletitle{HaWoR: World-Space Hand Motion Reconstruction from Egocentric Videos}.
\newblock \bibinfo{journal}{\emph{arXiv preprint arXiv:2501.02973}} (\bibinfo{year}{2025}).
\newblock


\bibitem[Zhang et~al\mbox{.}(2023)]%
        {controlnet:zhang2023adding}
\bibfield{author}{\bibinfo{person}{Lvmin Zhang}, \bibinfo{person}{Anyi Rao}, {and} \bibinfo{person}{Maneesh Agrawala}.} \bibinfo{year}{2023}\natexlab{}.
\newblock \bibinfo{title}{Adding Conditional Control to Text-to-Image Diffusion Models}.
\newblock


\bibitem[Zhao et~al\mbox{.}(2025)]%
        {Zhao_2025_CVPR}
\bibfield{author}{\bibinfo{person}{Zhuoran Zhao}, \bibinfo{person}{Linlin Yang}, \bibinfo{person}{Pengzhan Sun}, \bibinfo{person}{Pan Hui}, {and} \bibinfo{person}{Angela Yao}.} \bibinfo{year}{2025}\natexlab{}.
\newblock \showarticletitle{Analyzing the Synthetic-to-Real Domain Gap in 3D Hand Pose Estimation}. In \bibinfo{booktitle}{\emph{Proceedings of the Computer Vision and Pattern Recognition Conference (CVPR)}}. \bibinfo{pages}{12255--12265}.
\newblock


\bibitem[Zhou et~al\mbox{.}(2024)]%
        {zhou2024simple}
\bibfield{author}{\bibinfo{person}{Zhishan Zhou}, \bibinfo{person}{Shihao. zhou}, \bibinfo{person}{Zhi Lv}, \bibinfo{person}{Minqiang Zou}, \bibinfo{person}{Yao Tang}, {and} \bibinfo{person}{Jiajun Liang}.} \bibinfo{year}{2024}\natexlab{}.
\newblock \bibinfo{title}{A Simple Baseline for Efficient Hand Mesh Reconstruction}.
\newblock


\end{thebibliography}

\appendix

\newpage

This is the Appendix/Supplementary pages for paper ``Estimating Accurate Hand Pose in Camera Space with Vision Transformer''.


\section{Perspective Information Embedding}
\label{sec:pie}

We detailed the Perspective Information Embedding (PIE) module in \cref{sec:method-pie} of the main paper. Here we provide more implementation details. As we stated above, the PIE module aims to model the effect of perspective projection process and camera intrinsics.

\subsection{Perspective Information Encoding from Camera Rays}
\label{sec:pie-camera-ray-encoding}

\begin{figure}[!hbt]
    \centering
    \includegraphics[width=0.5\linewidth]{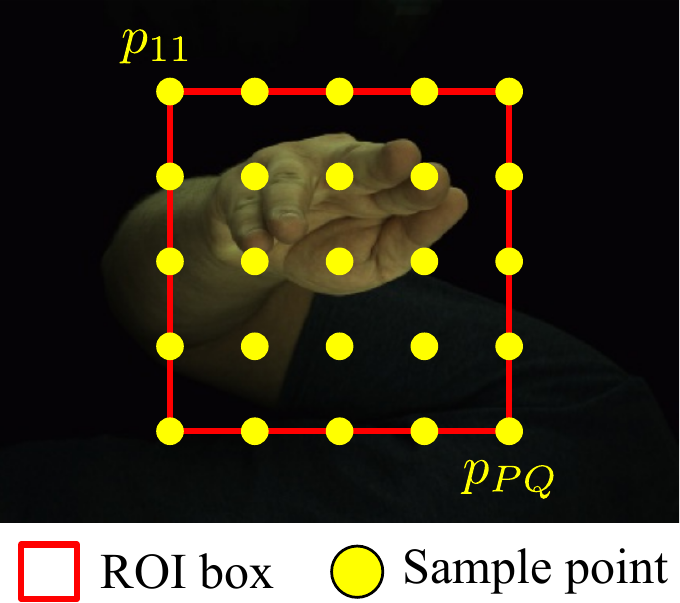}
    \caption{Sampling of Perspective Information Embedding (PIE) module.}
    \label{fig:pie-sample}
\end{figure}

Given the ROI of hands $\mathcal B=(x_1,y_1,x_2,y_2)$ in the image and camera intrinsics $K\in\mathbb R^{3\times 3}$, we sample dense set of points within the bouding box to represent the perspective information. Specifically, we uniformly sample $N=P\times Q$ points within the bounding box $\mathcal B$, denoted as $\{p_{ij}\in\mathbb R^2\}^N_{ij}$, as shown in \cref{fig:pie-sample}.

The sampled points are then projected to the normalized camera space using the camera intrinsics $K$. This process normalized the pixel coordinates from image space into camera space, which is formulated as:
\begin{equation}
    u_{ij} = K^{-1}\cdot [p_{ij};1], u_{ij}\in\mathbb R^3.
\end{equation}
$u_{ij}$ is the normalized camera coordinate of point $p_{ij}$ which can be seen as the intersection of the ray from camera center through pixel $p_{ij}$ with the normalized image plane at $Z=1$. Homogeneous coordinate is used here for clarity, which is different from the main paper.

To simulate the process of perspective projection, we cast rays from $u_{ij}$ to the camera center and denoted each ray with the normalized direction vector:
\begin{equation}
    \hat u_{ij} = \frac{-u_{ij}}{\Vert u_{ij}\Vert_2}.
\end{equation}
As we have constrainted the norm of $\hat u_{ij}$ to 1, we dropped the last element of coordinate for simplicity. Each simplified direction vector $(\hat u_{ij})_{xy}$ is concatenated into $F_\text{persp}$ as the perspective information:
\begin{equation}
    F_\text{persp} = \text{cat}((\hat u_{11})_{xy},\cdots,(\hat u_{PQ})_{xy})\in\mathbb R^{N\times 2}.
\end{equation}

The $F_\text{persp}$ is then flatten into $\mathbb R^{2N}$ and passed through the Perspective Embedder (which is a MLP) to obtain the perspective embedding vector $v_\text{persp}=\text{PIE}(F_\text{persp})\in\mathbb R^d$. We assert that this vector contains the perspective information of hand related region in the input image, which can be helpful for alleviating the coupling of local and global pose under perspective projection.

\subsection{Focal-Length Normalization via Camera Intrinsics}
\label{sec:pie-focal-normalization}

PIE explicitly incorporates camera calibration by mapping an image point $p=[u,v,1]^\top$ to its normalized camera coordinate:
\begin{equation}
    q=K^{-1}p
    =\begin{bmatrix}
        (u-c_x)/f_x & (v-c_y)/f_y & 1
    \end{bmatrix}^{\top}.
\end{equation}
Consider the same viewing ray observed by cameras with intrinsic matrices $K$ and $K'$. Its image coordinates satisfy $p=Kq$ and $p'=K'q$, respectively, and therefore
\begin{equation}
    K^{-1}p=K'^{-1}p'=q.
\end{equation}
For example, if the focal length and the corresponding image displacement from the principal point are both scaled by $\lambda$, then $f'_x=\lambda f_x$ and $u'-c'_x=\lambda(u-c_x)$, giving
\begin{equation}
    \frac{u'-c'_x}{f'_x}=\frac{u-c_x}{f_x}.
\end{equation}
The same relation holds for the vertical coordinate. Thus, when image projection and the supplied camera intrinsics vary consistently, PIE represents the sampled pixels as the same normalized camera rays and
explicitly accounts for focal-length differences. Therefore, given correctly calibrated intrinsics, PIE accounts for focal-length variations by encoding image points as normalized camera rays.


\section{Specification of Scale-Rotation Transformation}
\label{sec:transformation}
The scale-rotation transformation is isomorphic in image and 3D space, which we will detail their specification in each space here.

In the image space, the scale-rotation transformation is defined as (i) first scale the image coordinates with a scale factor $s$ around the center, and then (ii) rotate the scaled coordinates with a rotation matrix $R_2(\alpha)\in SO(2)$ around the same center. Formally, given a pixel coordinate $p\in\mathbb R^2$ in the image, the transformed pixel coordinate $p'$ is computed as:
\begin{equation}
    p' = R_2(\alpha)\cdot (p-c)\cdot s + c,
\end{equation}
where $c\in\mathbb R^2$ is the center of transformation.

In the 3D space, the scale-rotation transformation is defined as (i) shift the $z$-axis by multiplying $1/s$, and then (ii) rotate the scaled coordinates with a rotation matrix $R_z(\alpha)\in SO(3)$ around the $z$-axis. Formally, given a 3D coordinate $P\in\mathbb R^3$, the transformed 3D coordinate $P'$ is computed as:
\begin{equation}
    P'=R_z(\alpha)\cdot P_s, \quad P_s =
    \begin{bmatrix}
    P_x\\
    P_y\\
    P_z/s
    \end{bmatrix}.
\end{equation}

For any 3D point $P$, its 2D projection under intrinsics $(f_x,f_y,c_x,c_y)$ is denoted as pixel coordinate $p$. After applying the above scale-rotation transformation in both image and 3D space, the transformed 3D point $P'$ will be projected to the transformed pixel coordinate $p'$. This property shows the isomorphism of the scale-rotation transformation in image and 3D space. Following the ordinary image-space rotation used here, we
consider the case $f_x=f_y=f$ and let the transformation center coincide
with the camera principal point, i.e., $c=[c_x,c_y]^\top$. Here is a brief proof:

Given a 3D point $P$ and its projection $p$:
\begin{equation}
    p_x = f_x\frac{P_x}{P_z}+c_x, \quad
    p_y = f_y\frac{P_y}{P_z}+c_y.
\end{equation}
After applying the scale-rotation transformation in 3D space, we have:
\begin{equation}
    P' = R_z(\alpha)\cdot
    \begin{bmatrix}
    P_x\\
    P_y\\
    P_z/s
    \end{bmatrix} = R_z(\alpha)\cdot P_s.
\end{equation}
The projection of $P'$ is computed as:
\begin{equation}
   p'_x = f_x\frac{(R_z(\alpha)\cdot P_s)_x}{(R_z(\alpha)\cdot P_s)_z}+c_x, \quad
   p'_y = f_y\frac{(R_z(\alpha)\cdot P_s)_y}{(R_z(\alpha)\cdot P_s)_z}+c_y.
\end{equation}
On the other hand, applying the scale-rotation transformation in image space, we have:
\begin{equation}
    p' = R_2(\alpha)\cdot (p-c)\cdot s + c.
\end{equation}
Writing $R_2(\alpha)=[R_{ij}]$ and substituting $p$ into the above equation, we have:
\begin{equation}
\begin{aligned}
    p'_x = &\,R_{11}\left(f_x\frac{P_x}{P_z}+c_x - c_x\right)s \\ &+ R_{12}\left(f_y\frac{P_y}{P_z}+c_y - c_y\right)s + c_x, \\
    p'_y = &\,R_{21}\left(f_x\frac{P_x}{P_z}+c_x - c_x\right)s \\ &+ R_{22}\left(f_y\frac{P_y}{P_z}+c_y - c_y\right)s + c_y.
\end{aligned}
\end{equation}
With out loss of generality, we list the derivation for $p'_x$ here. Using $f_x=f_y=f$, the $x$ coordinate can be simplified as:
\begin{equation}
\begin{aligned}
    p'_x &= R_{11}(f\frac{P_x}{P_z})s + R_{12}(f\frac{P_y}{P_z})s + c_x \\
    &= f\frac{R_{11}P_x s + R_{12}P_y s}{P_z} + c_x \\
    &= f\frac{(R_z(\alpha)\cdot P_s)_x}{(R_z(\alpha)\cdot P_s)_z} + c_x \\
    &= f\frac{P'_x}{P'_z} + c_x.
\end{aligned}
\end{equation}
The derivation for $p'_y$ is similar. Thus we prove that the scale-rotation transformation is isomorphic in image and 3D space.

Specifically, since the input to $sr_\text{pose}$ consists of the joint rotations $\theta$, shape parameters $\beta$, and root joint translation $t$, we define the 3D space scale-rotation operation as applying a rotation to the root joint orientation and translation, and a $z$-axis scaling of the root joint translation.

\begin{figure}[H]
    \centering
    \includegraphics[width=\linewidth]{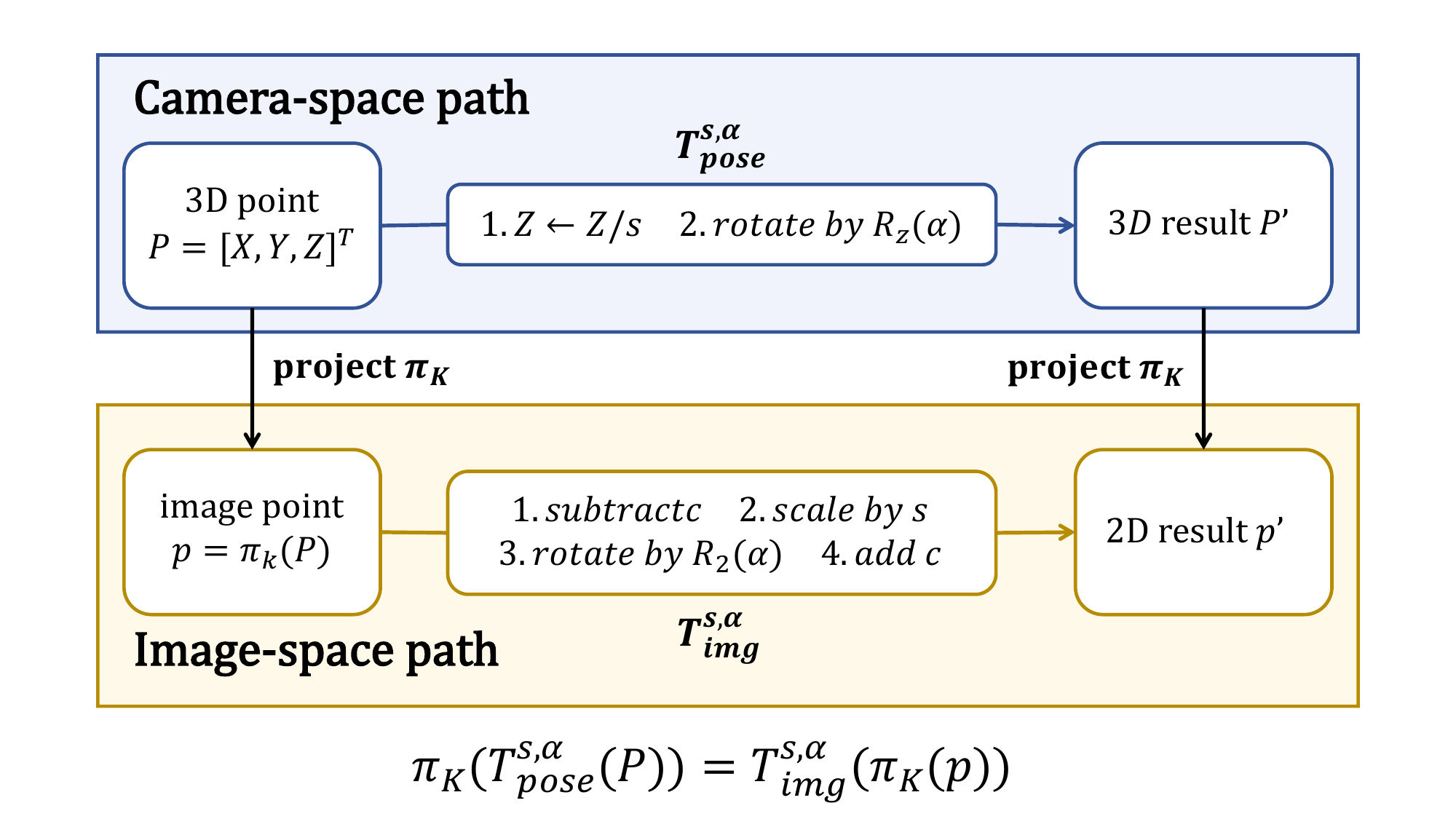}
    \caption{Scale-rotation correspondence in camera and image spaces.}
    \label{fig:tis-process}
\end{figure}

\noindent\textbf{Step-by-Step Example.}
Consider $P=[20,10,500]^\top$ mm, $f_x=f_y=600$ px,
$c=[320,240]^\top$ px, $s=1.2$, and $\alpha=30^\circ$. As illustrated in
\cref{fig:tis-process}, the camera-space path first scales the depth from
$500$ mm to $500/1.2=416.7$ mm and then rotates $(20,10)$ to approximately
$(12.3,18.7)$ mm; projecting the resulting point gives
$p'=[337.7,266.9]^\top$ px. In the image-space path, the original projection
$p=[344,252]^\top$ px is centered as $p-c=[24,12]^\top$ px, scaled to
$[28.8,14.4]^\top$ px, rotated to $[17.7,26.9]^\top$ px, and translated
back by $c$, producing the same $p'$.

TIS uses this correspondence to regularize the learned latent transformation during training.


\begin{figure}[H]
    \centering
    \includegraphics[width=0.3\linewidth]{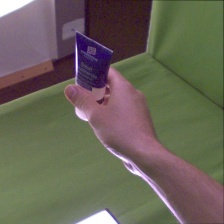}\,
    \includegraphics[width=0.3\linewidth]{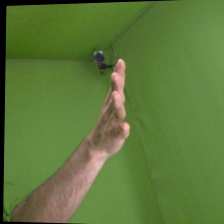}\,
    \includegraphics[width=0.3\linewidth]{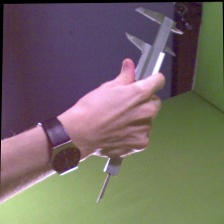}
    \caption{Images from FreiHAND dataset.}
    \label{fig:freihand-images}
\end{figure}

\section{About FreiHAND Dataset Results}
\label{sec:freihand-dataset-analysis}

FreiHAND is a large-scale, single-image 3D hand pose and shape estimation benchmark. It contains over 130,000 training frames, each providing accurate ground truth 3D hand joint annotations (21 keypoints) and MANO shape parameters. The dataset was captured using a multi-view green-screen setup, enabling easy background augmentation and robust training. It has become a standard benchmark for evaluating 3D hand reconstruction from a single RGB image.

As shown in \cref{fig:freihand-images}, we note that the images in the FreiHAND dataset have already been normalized. The normalization method involves re-projecting the image via a perspective transformation, which aligns the camera's optical axis with the center of the hand bounding box, following resizing the image to $224\times224$.

After this transformation, the resulting image is analogous to one captured by a telephoto lens, exhibiting diminished perspective distortion and more closely approximating an orthographic projection. Moreover, in telephoto scenarios, estimating camera intrinsic parameters is more challenging. This is because variations in object distance produce only subtle changes in projection size. This effect, compounded by factors such as camera shake and defocus during capture, typically exacerbates errors in depth perception. For a machine learning model, this means it is highly prone to overfitting to the dataset's specific characteristics, rendering it unable to accurately estimate the 3D hand pose in camera space.

\begin{figure}[H]
    \centering
    \includegraphics[width=\linewidth]{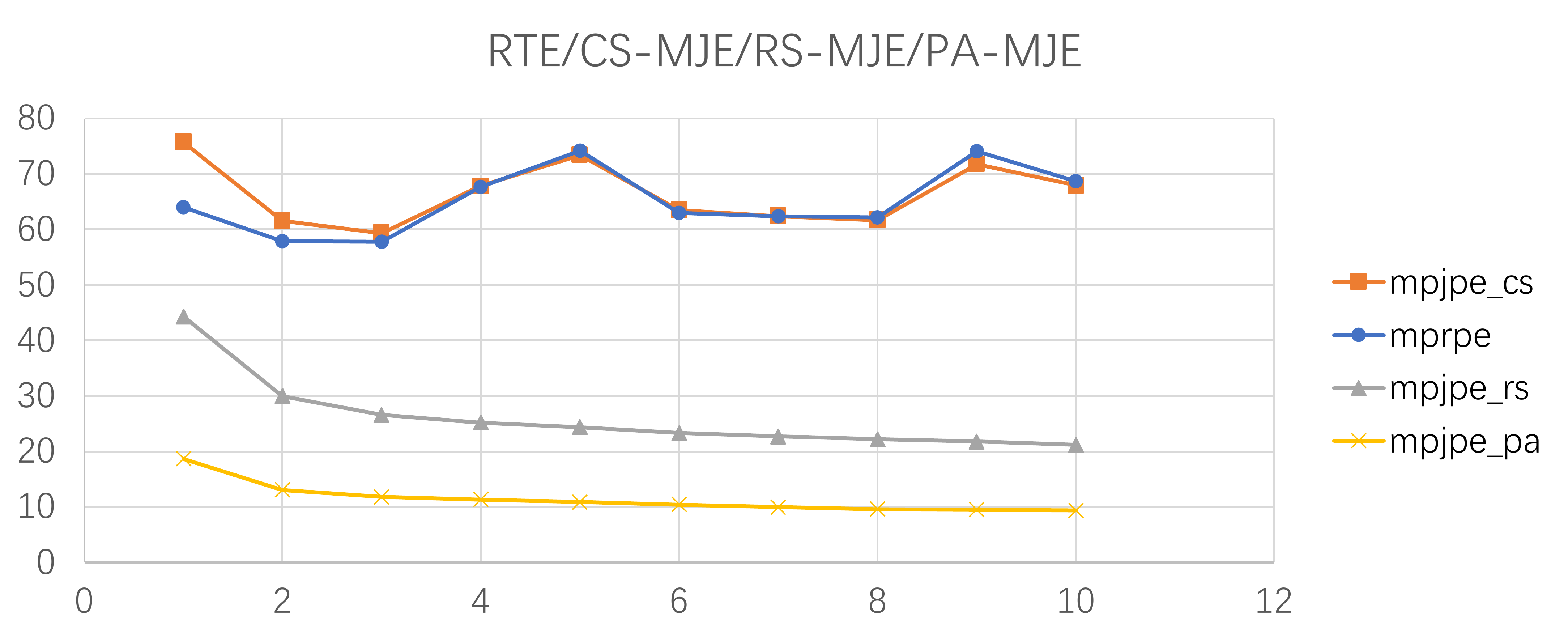}
    \caption{Per-epoch metrices on test set of FreiHAND.}
    \label{tab:freihand-epoch-metrices}
\end{figure}

\begin{figure}[H]
    \centering
    \includegraphics[width=\linewidth]{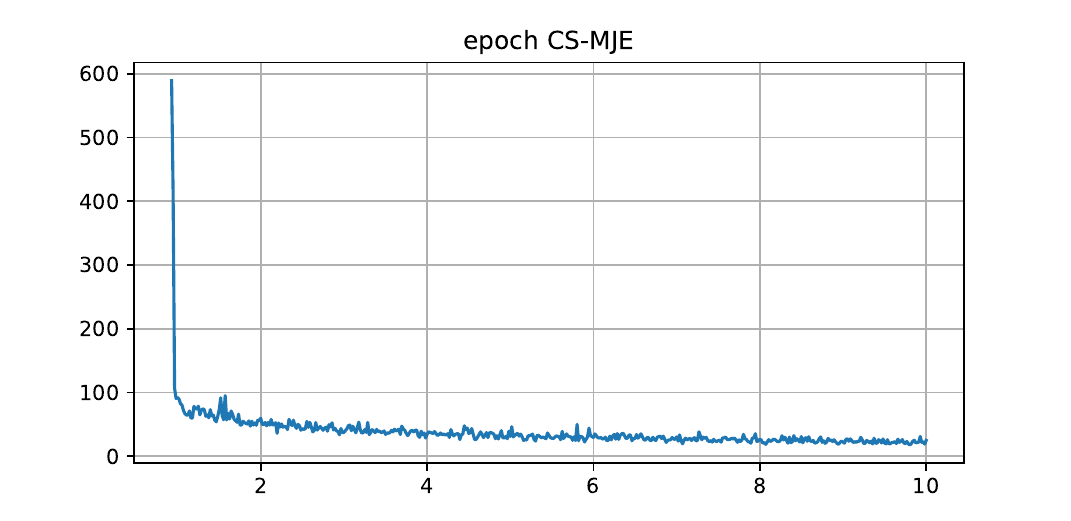}
    \caption{Per-step CS-MJE during the training.}
    \label{tab:freihand-running-metrices}
\end{figure}

We trained our model on FreiHAND and plotted the CS-MJE, RTE, RS-MJE and PA-MJE of test set for each epoch, the results are shown in \cref{tab:freihand-epoch-metrices}. We observed that the metrics reflecting local hand pose estimation (RS-MJE and PA-MJE) decreased and converged to reasonable levels during training. However, the camera-space hand pose estimation metrics (CS-MJE and RTE) continuously oscillated at a high error level. By comparison, we can determine that the main source of this error is root joint localization, as the RTE metric remained persistently high. Conversely, \cref{tab:freihand-running-metrices} shows that the CS-MJE on the training set does decrease, converging to below 30 mm. This divergence from the test set performance confirms that the model has overfit on the global pose estimation task, which validates our theory.

\section{Additional Quantitative Evaluations}
\label{sec:additional-quantitative-evaluations}

\subsection{PnP Error Analysis}
\label{sec:pnp-error-analysis}

We analyze the PnP post-processing used to lift HaMeR predictions into camera space on the HO3D evaluation set. Root-PnP recovers the camera-space root position by estimating only a global translation $(t_x,t_y,t_z)$, while retaining HaMeR's predicted global orientation and MANO-derived hand scale.
The translation is obtained by minimizing the 2D reprojection error over all 21 joint correspondences. In contrast, Full-PnP jointly estimates global rotation and translation. Both variants use HaMeR-predicted 2D
correspondences, HO3D ground-truth camera intrinsics, and GT-derived hand bounding boxes.

\begin{table}[!hbt]
    \small
    \centering
    \setlength{\tabcolsep}{3pt}
    \caption{PnP post-processing results on the HO3D. GT replaces HaMeR-predicted 2D correspondences with ground-truth 2D joints. All errors are in millimeters.}
    \label{tab:pnp-error-analysis}
    \begin{tabular}{lcccc}
        \toprule
        Method & CS-MJE$\downarrow$ & RS-MJE$\downarrow$
        & PA-MJE$\downarrow$ & RTE$\downarrow$ \\
        \midrule
        HaMeR-H + Root-PnP          & 44.15 & 14.50 & 7.17  & 44.19 \\
        HaMeR-H + Full-PnP          & 33.83 & 19.11 & 7.17  & 36.00 \\
        HaMeR-H + Full-PnP + GT     & 27.55 & 11.33 & 7.17  & 26.68 \\
        \midrule
        HaMeR-B + Root-PnP          & 63.52 & 35.37 & 11.06 & 56.77 \\
        HaMeR-B + Full-PnP          & 55.60 & 35.84 & 11.06 & 48.51 \\
        HaMeR-B + Full-PnP + GT     & 53.37 & 16.74 & 11.06 & 51.14 \\
        \bottomrule
    \end{tabular}
\end{table}

As shown in \cref{tab:pnp-error-analysis}, HaMeR-H Root-PnP obtains 44.15 mm CS-MJE and 44.19 mm RTE, while its wrist-aligned RS-MJE is 14.50 mm. The difference shows that its camera-space error is dominated by global placement. Full-PnP reduces CS-MJE to 33.83 mm but increases RS-MJE to 19.11 mm, indicating that fitting rotation to predicted 2D joints can improve global placement while degrading root-relative orientation. HaMeR-B remains less accurate under both variants. Replacing predicted 2D correspondences with GT 2D joints substantially reduces RS-MJE for both backbones, confirming that 2D keypoint errors are an important, but not exclusive, source of the remaining error.

\begin{figure}[H]
    \centering
    \includegraphics[width=0.9\linewidth]{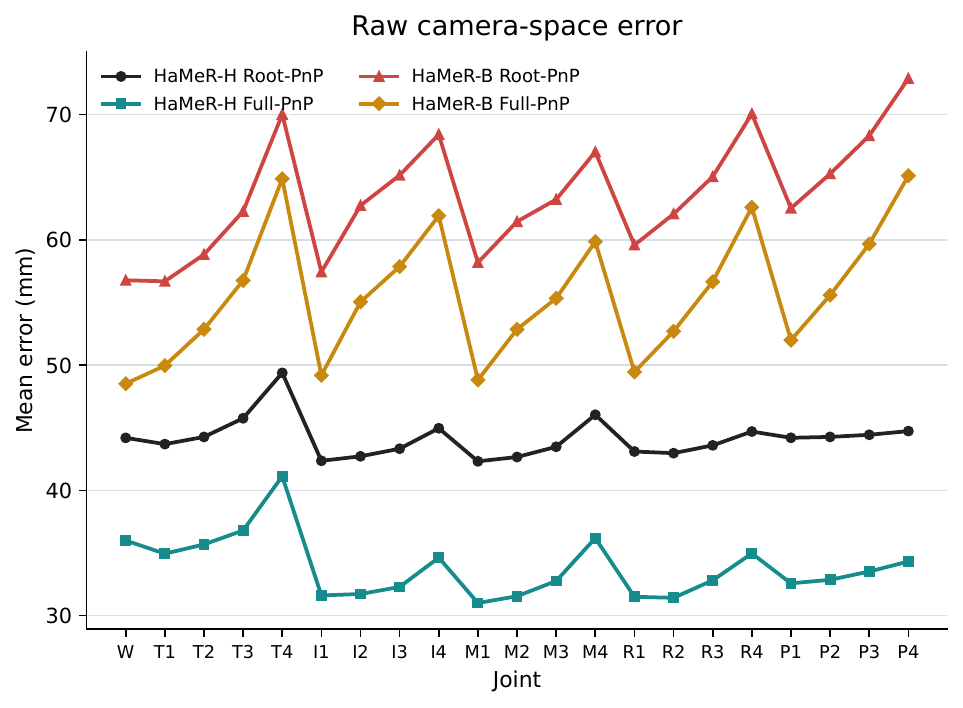}
    \caption{Per-joint camera-space PnP errors. W denotes the wrist; T, I, M, R, and P denote the five fingers; levels 1–4 run from the palm-side joint to the fingertip.}
    \label{fig:pnp-cs}
\end{figure}

\begin{figure}[H]
    \centering
    \includegraphics[width=0.9\linewidth]{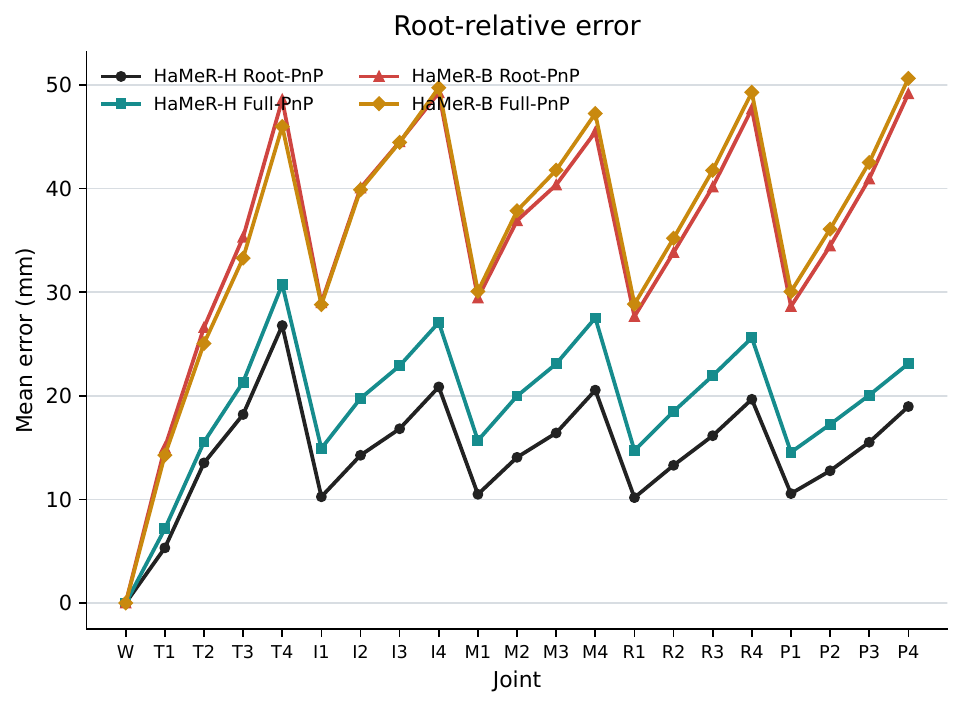}
    \caption{Per-joint root-relative PnP errors.}
    \label{fig:pnp-rs}
\end{figure}

The per-joint results in \cref{fig:pnp-rs} show a consistent spatial pattern. Under Root-PnP, the mean root-relative error across the five fingers increases from 9.36 mm at level 1 to 21.36 mm at the fingertips for HaMeR-H, and from 25.94 to 48.02 mm for HaMeR-B. Translation-only fitting therefore cannot remove errors that accumulate along the kinematic chains.

\subsection{Robustness Analysis}
\label{sec:robustness-analysis}

\noindent\textbf{Bounding-box perturbations.}
We separately evaluate center and scale errors on the HO3D evaluation set. For center perturbations, we fix the bounding-box scale to 1 and shift its center relative to the box size. Random selects a uniformly distributed
direction for each sample, radial follows the direction from the camera principal point to the original box center, and tangential is perpendicular to the radial direction. Compared with the clean CS-MJE of 46.36 mm, offsets of 2.5\% and 5\% cause only modest degradation, while a 20\% offset yields 48.95, 50.92, and 54.19 mm for radial, random, and tangential shifts, respectively. For scale perturbations, we keep the center fixed and vary the scale from 0.8 to 1.2. CS-MJE is lowest at scale 1.0 and reaches 47.55 and 47.47 mm at scales 0.8 and 1.2, showing a modest and approximately symmetric degradation as the scale departs from its original value.

\begin{figure}[H]
    \centering
    \includegraphics[width=0.9\linewidth]{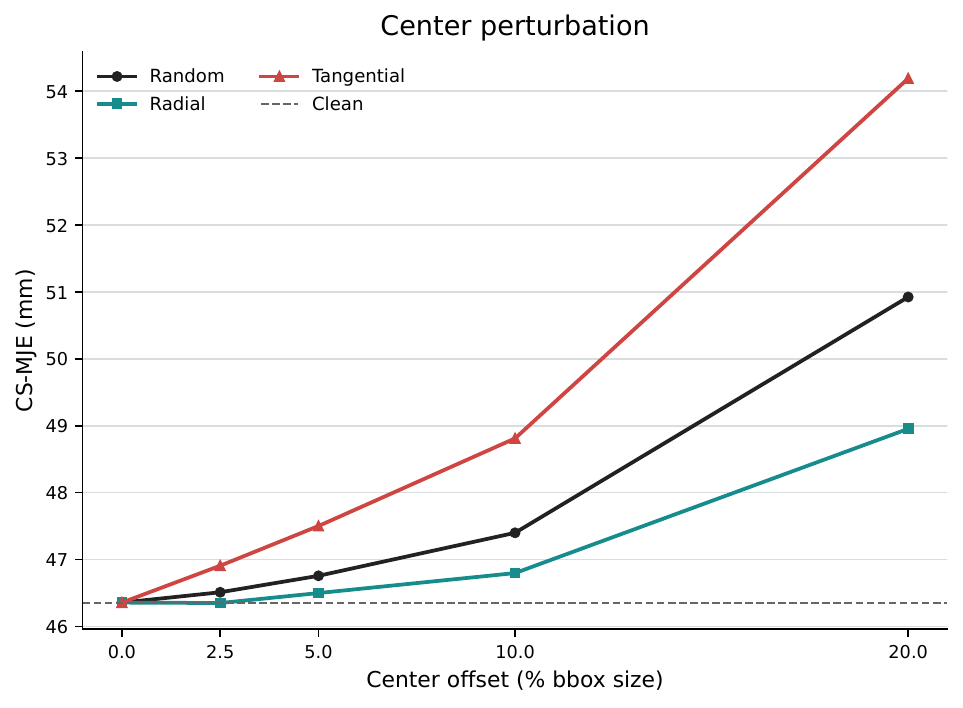}
    \caption{Bounding-box center robustness on HO3D.}
    \label{fig:bbox-center}
\end{figure}

\begin{figure}[H]
    \centering
    \includegraphics[width=0.9\linewidth]{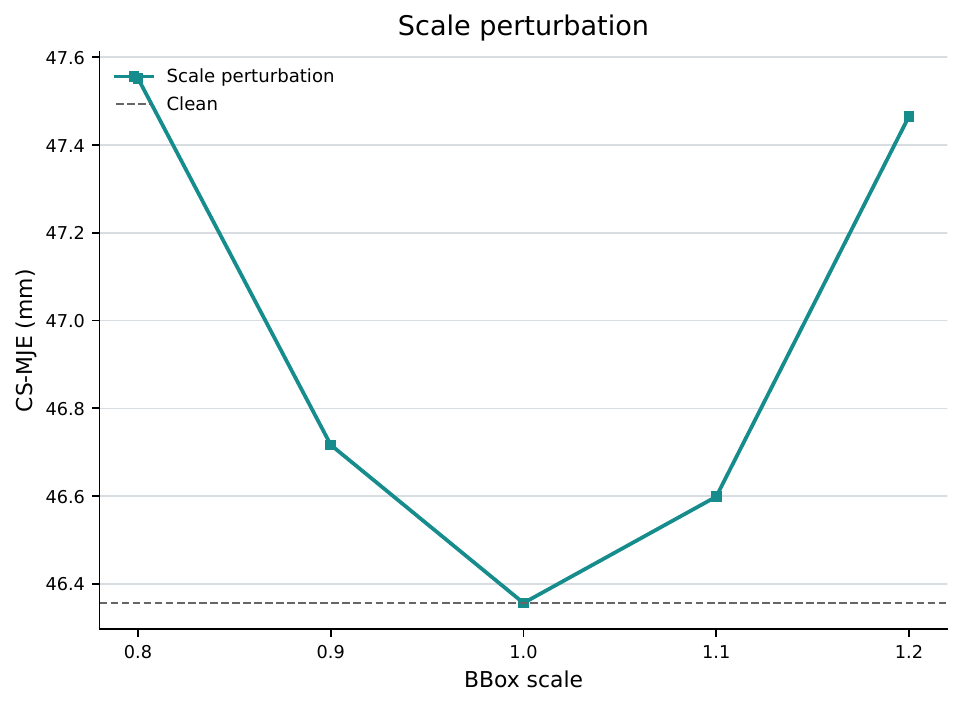}
    \caption{Bounding-box scale robustness on HO3D.}
    \label{fig:bbox-scale}
\end{figure}

\noindent\textbf{Calibration-error sensitivity.}
We keep the image crop fixed and perturb only the camera intrinsics supplied to PIE. Focal-length errors within $\pm5\%$ change CS-MJE by less than 1\%, while focal scales of 0.8 and 1.2 increase CS-MJE to 48.08 and 49.07 mm,
respectively. A 5\% principal-point shift increases CS-MJE to between 48.71 and 50.88 mm, depending on direction. Calibration errors therefore increase camera-space joint and wrist-translation errors, with larger
perturbations causing greater degradation.

\begin{figure}[H]
    \centering
    \includegraphics[width=0.9\linewidth]{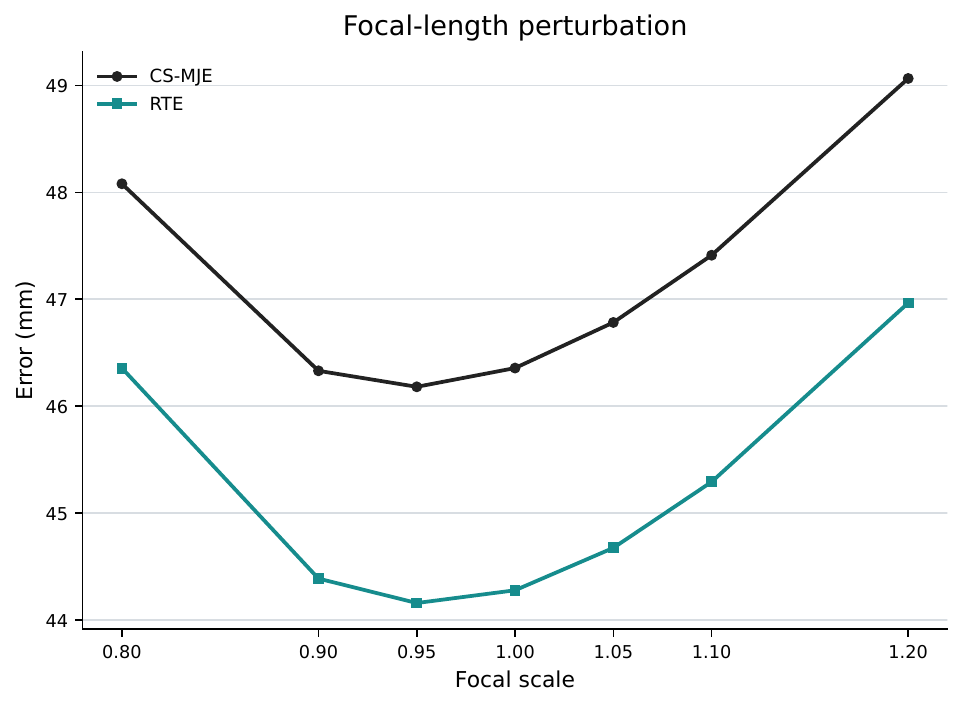}
    \caption{Focal-length robustness on HO3D.}
    \label{fig:focal-sensitivity}
\end{figure}

\begin{figure}[H]
    \centering
    \includegraphics[width=0.9\linewidth]{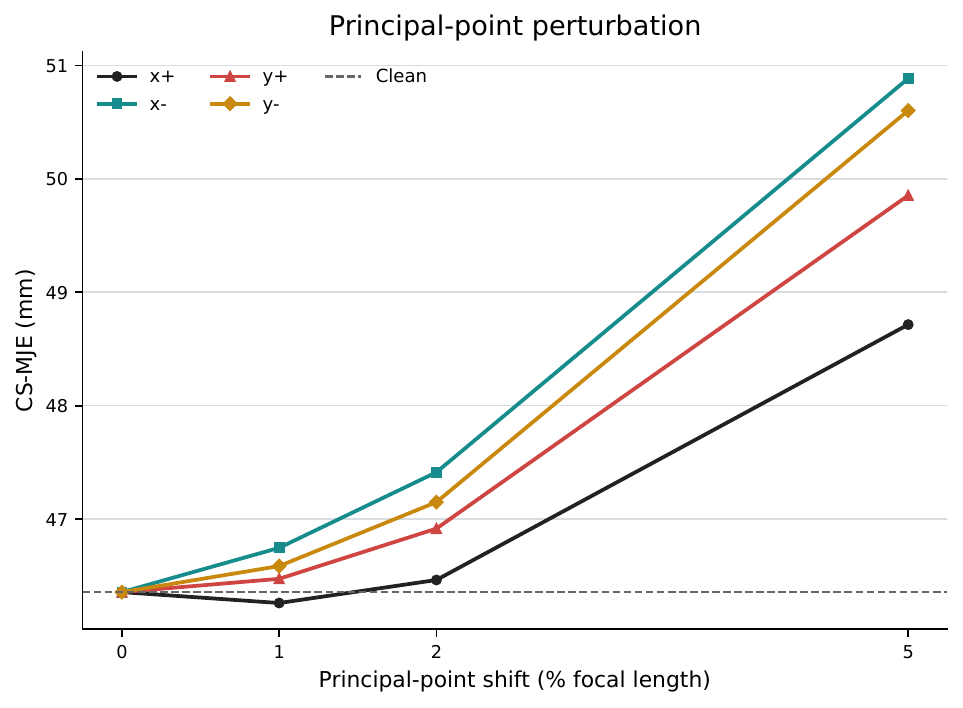}
    \caption{Principal-point robustness on HO3D.}
    \label{fig:principal-sensitivity}
\end{figure}

\subsection{Additional Ablation on DexYCB}
\label{sec:dexycb-ablation}

We additionally evaluate the main components on DexYCB. The results are summarized in \cref{tab:dexycb-ablation}.

\begin{table}[!hbt]
    \centering
    \caption{DexYCB ablation results (mm).}
    \label{tab:dexycb-ablation}
    \begin{tabular}{lcccc}
        \toprule
        Method & CS-MJE$\downarrow$ & RS-MJE$\downarrow$
        & PA-MJE$\downarrow$ & RTE$\downarrow$ \\
        \midrule
        observ.          & 51.3 & 18.9 & 7.2 & 50.8 \\
        PIE              & 51.8 & \textbf{18.2} & 7.0 & 51.5 \\
        PIE + TIS        & \textbf{50.1} & 19.4 & \textbf{6.9}
                         & \textbf{49.5} \\
        \bottomrule
    \end{tabular}
\end{table}

Compared with the observation baseline, PIE reduces RS-MJE from 18.9 to 18.2 mm and PA-MJE from 7.2 to 7.0 mm, improving local-pose accuracy. Adding TIS reduces CS-MJE from 51.8 to 50.1 mm and RTE from 51.5 to 49.5 mm, while RS-MJE increases from 18.2 to 19.4 mm. The DexYCB results therefore support a camera-space benefit from TIS while revealing a trade-off in root-relative pose accuracy.


\section{Training details}
\label{sec:train-detail}
\begin{table}[!hbt]
    \centering
    \caption{Details of spatial training phase for DexYCB and HO3D.}
    \label{tab:spatial-train-detail}
    \begin{tabular}{ll}
        \toprule
        Property & Value \\
        \midrule
        epoch & 30 \\
        batch size & 32 \\
        learning rate & $10^{-4}$ \\
        optimizer & AdamW \\
        learning rate scheduler & constant \\
        \bottomrule
    \end{tabular}
\end{table}

\begin{table}[!hbt]
    \centering
    \caption{Details of temporal training phase.}
    \label{tab:temporal-train-detail}
    \begin{tabular}{lll}
        \toprule
        Dataset & Property & Value \\
        \midrule
        \multirow{5}{*}{HO3D} & epoch & 30 \\
        & batch size & 36 \\
        & learning rate & $10^{-4}$ \\
        & frame length & 7 \\
        & optimizer & AdamW \\
        & learning rate scheduler & constant \\

        \midrule
        \multirow{5}{*}{DexYCB} & epoch & 10 \\
        & batch size & 32 \\
        & learning rate & $10^{-4}$ \\
        & frame length & 7 \\
        & optimizer & AdamW \\
        & learning rate scheduler & constant \\

        \midrule
        \multirow{5}{*}{MDT} & epoch & 1 \\
        & batch size & 45 \\
        & learning rate & $10^{-4}$ \\
        & frame length & 7 \\
        & optimizer & AdamW \\
        & learning rate scheduler & constant \\
        \bottomrule
    \end{tabular}
\end{table}


\section{Qualitative Results}
\label{sec:qualitative-supp}

\subsection{More cases}
\begin{figure*}
    \centering
    \includegraphics[width=\linewidth]{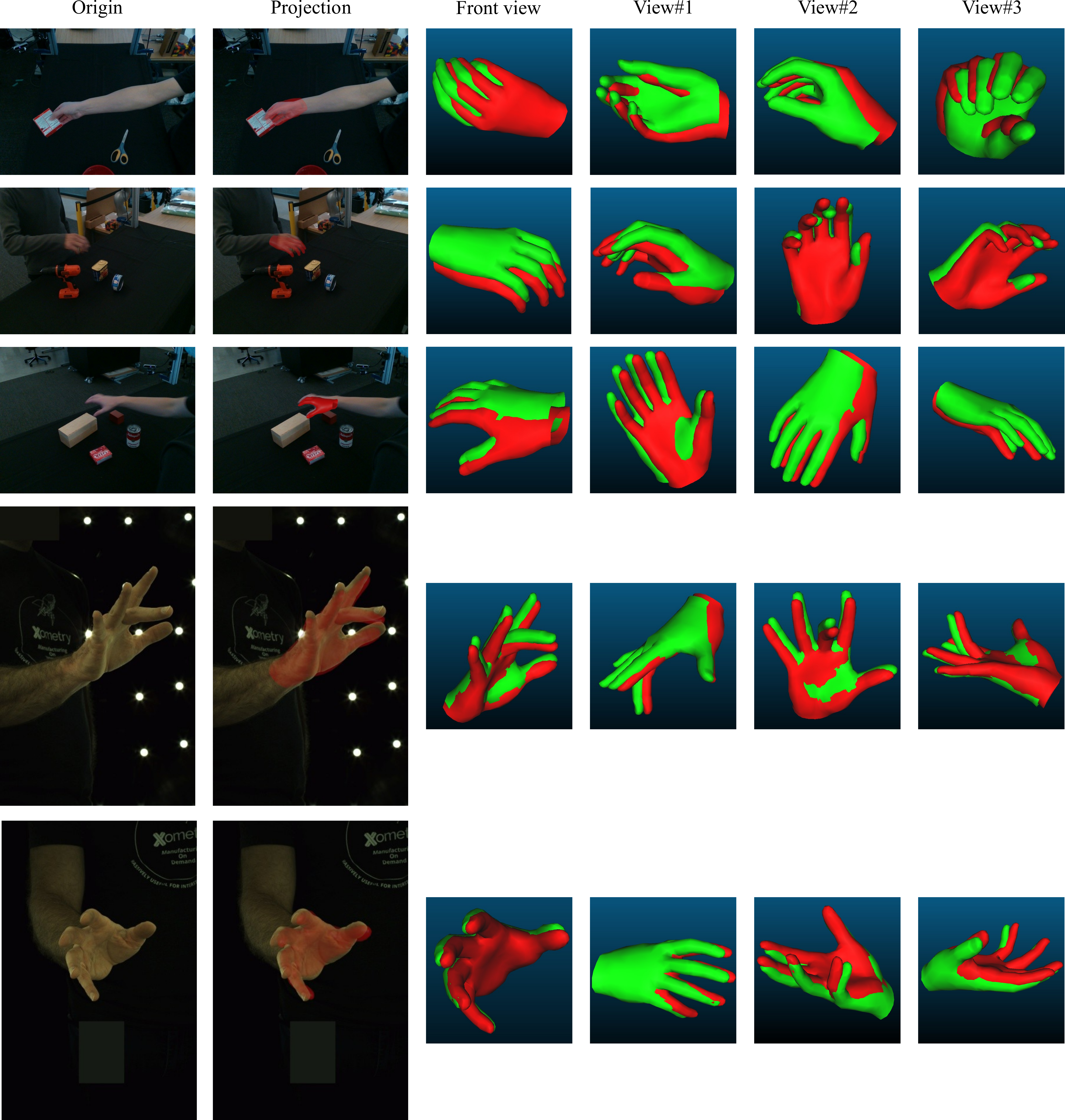}
    \caption{Qualitative results on DexYCB and InterHand2.6M.}
    \label{fig:qualitative-supp}
\end{figure*}

Please refer to \cref{fig:qualitative-supp}.

\subsection{Root alignment cases}

\begin{figure*}[t]
    \centering
    \includegraphics[width=0.8\linewidth]{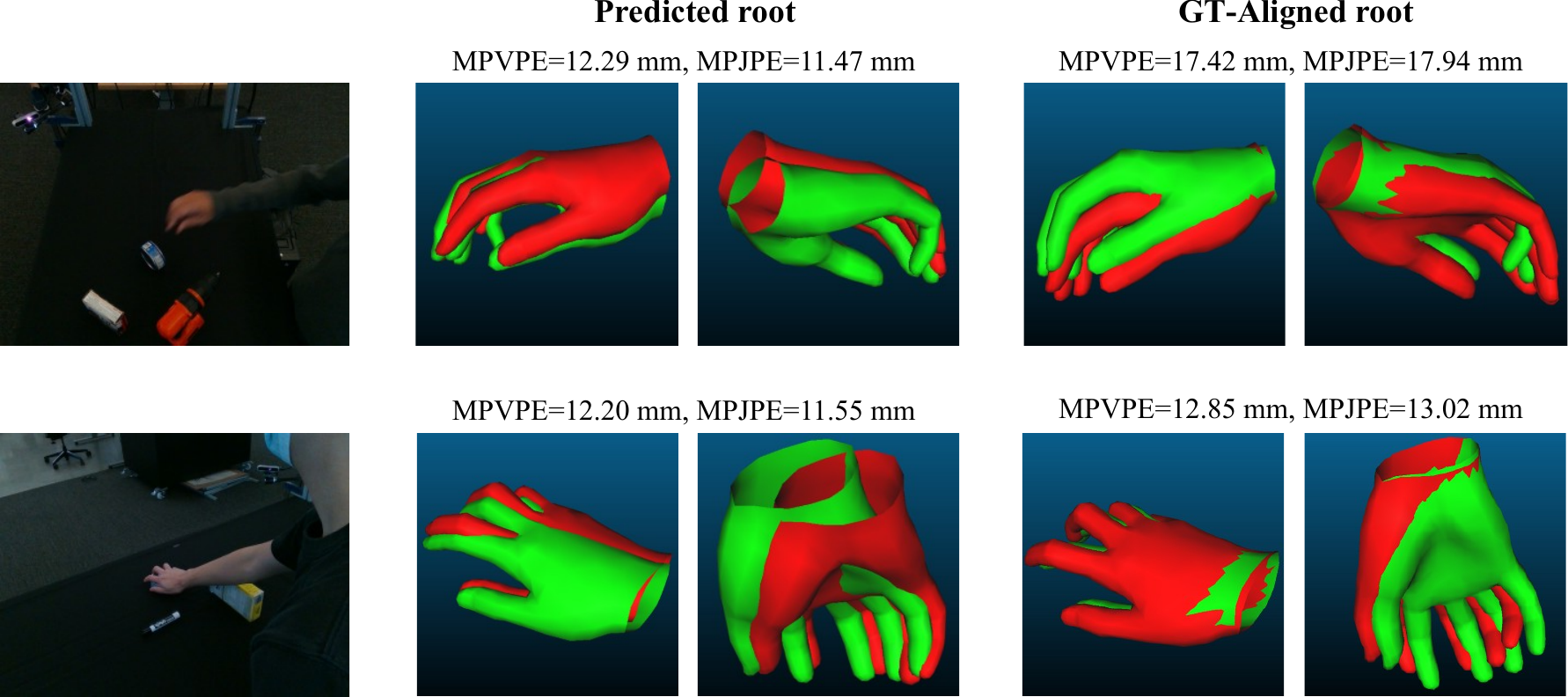}
    \caption{DexYCB dataset. Qualitative and quantitative results for alignment choices. The ``Predicted'' use the wrist position predicted by our model and ``GT-Aligned'' use the ground-truth wrist position extracted from the annotations.}
    \label{fig:ra-supp}
\end{figure*}

\begin{figure*}[t]
    \centering
    \includegraphics[width=0.8\linewidth]{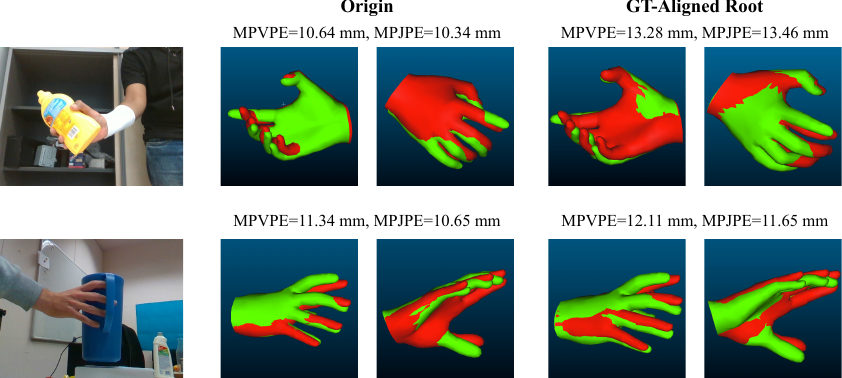}
    \caption{HO3D dataset. Qualitative and quantitative results for alignment choices. The ``Predicted'' use the wrist position predicted by our model and ``GT-Aligned'' use the ground-truth wrist position extracted from the annotations.}
    \label{fig:ra-ho3d-supp}
\end{figure*}

\cref{fig:ra-supp} and \cref{fig:ra-ho3d-supp} show the difference between two visualization routines on DexYCB and HO3D. In the ``Predicted root'' column, we render the predicted pose (red) and true pose (green) in the camera-space without any modifications to the predicted wrist. Whereas in local pose cases, the wrist position is manually modified to the ground-truth wrist position, as rendered in ``GT-Aligned root'' column.

Directly utilizing the estimated root joint position for rendering imposes stricter requirements on pose estimation accuracy, as the model must accurately localize the hand root joint within a vast free space to produce sufficiently accurate rendering results. In contrast, the root-joint alignment operation completely eliminates the difficulties arising from such high degrees of freedom, allowing the model to focus solely on local accuracy. Nevertheless, even when facing such challenges, our model can achieve higher pre-alignment accuracy than post-alignment accuracy in certain scenarios. This is because when root joint estimation is sufficiently accurate, the perspective coupling effect becomes pronounced. Our model pursues overall pose estimation accuracy in camera space (CS-MJE), whereas root joint alignment inadvertently increases error accumulation at distal joints. This characteristic enables our model to achieve more accurate estimation precision for joints that are more frequently involved in interactions, such as fingertips. Quantitative results on DexYCB for every joint are shown in \cref{tab:root_align_5_dex} and \cref{tab:root_align_9_dex}, corresponding to row 1 and 2 in \cref{fig:ra-supp} respectively.Quantitative results on HO3D for every joint are shown in \cref{tab:root_align_1_ho3d} and \cref{tab:root_align_2_ho3d}, corresponding to row 1 and 2 in \cref{fig:ra-ho3d-supp} respectively.

\subsection{Failure cases}
\begin{figure*}
    \centering
    \includegraphics[width=0.5\linewidth]{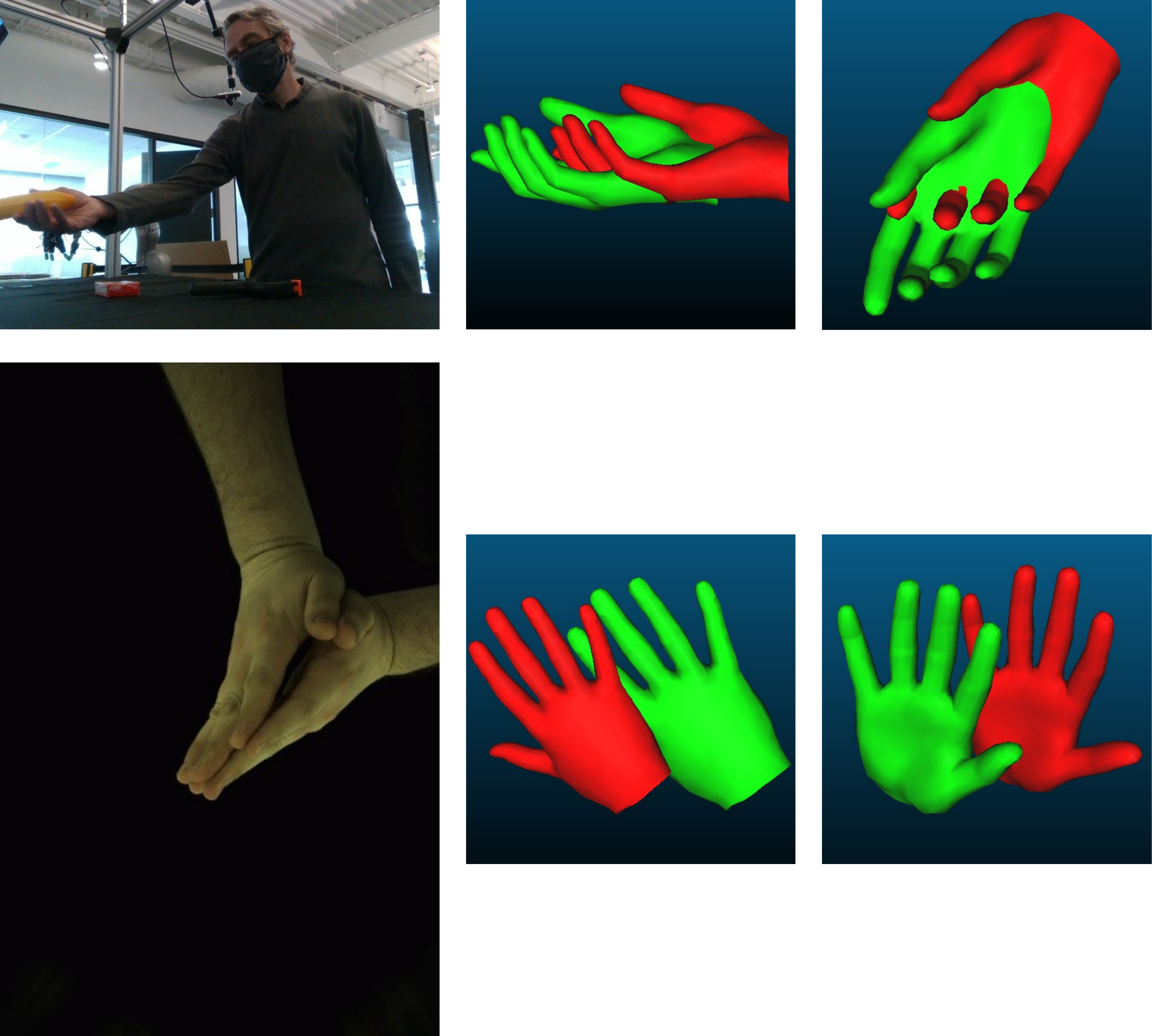}
    \caption{Failure cases of DexYCB and InterHand2.6M.}
    \label{fig:qualitative-fail-supp}
\end{figure*}

Failure cases are shown in \cref{fig:qualitative-fail-supp}.

\section{Jacobian Analysis of Perspective Projection and Perspective Coupling Effect}
\label{app:jacobian-coupling}

\subsection{Pinhole Camera Model}
\label{subsec:app-pinhole-camera}

The pinhole camera model describes the projection of a 3D spatial point $P_{\text{cam}} = [X, Y, Z]^\top$ to a 2D image point $p = [u, v]^\top$. This model is characterized by the camera intrinsic matrix $K \in \mathbb{R}^{3 \times 3}$:
\begin{equation}
K = \begin{bmatrix}
f_x & 0 & c_x \\
0 & f_y & c_y \\
0 & 0 & 1
\end{bmatrix},
\end{equation}
where $f_x, f_y$ are the focal lengths in the $x,y$ directions (in pixels), and $c_x, c_y$ are the principal point coordinates (image center). The projection process is:
\begin{equation}
\label{eq:app-perspective-projection}
p = \begin{bmatrix} u \\ v \end{bmatrix} = \pi(P_{\text{cam}}) = \begin{bmatrix}
f_x \frac{X}{Z} + c_x \\[4pt]
f_y \frac{Y}{Z} + c_y
\end{bmatrix}.
\end{equation}

The projection equation \cref{eq:app-perspective-projection} reveals the fundamental property of perspective projection: depth $Z$ appears in the denominator, causing the foreshortening effect---objects closer to the camera appear larger in the image.

\subsection{Complete Jacobian Matrix of Pixel Coordinates}
\label{subsec:app-full-jacobian}

To fully characterize the perspective coupling effect, we compute the Jacobian matrix of pixel coordinates $p_i = [u_i, v_i]^\top$ with respect to the six-dimensional vector $[J_{\text{rel},i}^\top, t^\top]^\top$. This matrix comprehensively describes the combined effect of local pose changes $J_{\text{rel},i}$ and root joint position changes $t$ on the projection.

Consider the $i$-th joint point in camera space. The joint coordinates are determined by both local pose $J_{\text{rel},i}$ and global translation $t$:
\begin{equation}
J_{\text{cam},i} = J_{\text{rel},i} + t,
\end{equation}
with its projection being $p_i = \pi(J_{\text{cam},i}) = [u_i, v_i]^\top$, where:
\begin{equation}
u_i = f_x \frac{X_i}{Z_i} + c_x, \quad v_i = f_y \frac{Y_i}{Z_i} + c_y,
\end{equation}
and $X_i = J_{\text{rel},i,x} + t_x$, $Y_i = J_{\text{rel},i,y} + t_y$, $Z_i = J_{\text{rel},i,z} + t_z$.

Computing partial derivatives of pixel coordinates with respect to $[J_{\text{rel},i}^\top, t^\top]^\top$ yields the complete $2 \times 6$ Jacobian matrix:
\begin{equation}
\label{eq:app-full-jacobian}
\frac{\partial p_i}{\partial [J_{\text{rel},i}^\top, t^\top]^\top} = \begin{bmatrix}
\frac{f_x}{Z_i} & 0 & -\frac{f_x X_i}{Z_i^2} & \frac{f_x}{Z_i} & 0 & -\frac{f_x X_i}{Z_i^2} \\[4pt]
0 & \frac{f_y}{Z_i} & -\frac{f_y Y_i}{Z_i^2} & 0 & \frac{f_y}{Z_i} & -\frac{f_y Y_i}{Z_i^2}
\end{bmatrix}.
\end{equation}

The Jacobian matrix has an important property: the corresponding components with respect to $J_{\text{rel},i}$ and $t$ have identical partial derivatives, i.e., $\partial u_i/\partial J_{\text{rel},i,x} = \partial u_i/\partial t_x$, $\partial u_i/\partial J_{\text{rel},i,z} = \partial u_i/\partial t_z$, etc. This symmetry arises from the \textbf{additive structure} of camera coordinates $J_{\text{cam},i} = J_{\text{rel},i} + t$---local pose changes and root joint translation changes have identical contributions to camera coordinates, and thus have identical effects on projected coordinates $p_i$.

\subsection{Depth Ambiguity Analysis via Jacobian Matrix}
\label{subsec:app-depth-ambiguity-jacobian}

Based on the complete Jacobian matrix, this section analyzes the specific manifestation of depth ambiguity in camera-space hand pose estimation from the perspective of matrix symmetry. Depth ambiguity is an inherent difficulty in monocular vision: the absolute depth of an object cannot be uniquely determined from a single view.

Consider a spatial point $P_{\text{cam}} = [X, Y, Z]^\top$ and its scaled version along the line of sight $P'_{\text{cam}} = \lambda P_{\text{cam}} = [\lambda X, \lambda Y, \lambda Z]^\top$, where $\lambda > 0$. According to the projection equation \cref{eq:app-perspective-projection}:
\begin{equation}
p = \pi(P_{\text{cam}}) = \begin{bmatrix} f_x \frac{X}{Z} + c_x \\ f_y \frac{Y}{Z} + c_y \end{bmatrix} = \pi(P'_{\text{cam}}) = p'.
\end{equation}
Therefore, $P_{\text{cam}}$ and $P'_{\text{cam}}$ produce exactly the same projection $p$ in the image, making it impossible to distinguish whether the object is at depth $Z$ or $\lambda Z$ from a single image.

From the perspective of the Jacobian matrix, depth ambiguity is manifested in equation \cref{eq:app-full-jacobian} as the symmetry of the additive structure. The complete $2 \times 6$ Jacobian matrix shows complete symmetry between local pose $J_{\text{rel},i}$ and global translation $t$ in terms of projection change rates. This symmetry stems from the additivity of camera coordinates $J_{\text{cam},i} = J_{\text{rel},i} + t$: the contributions of $J_{\text{rel},i}$ and $t$ to $J_{\text{cam},i}$ are completely equivalent, and thus their effects on projection are also equivalent. This means that mathematically, it is impossible to distinguish between ``local scale changes'' and ``root joint translation changes'' from a monocular image---both produce exactly the same effects on projection change rates.

This symmetry directly leads to the ill-posedness of depth estimation: the same projection change $\Delta p_i$ may be caused by local pose changes $\Delta J_{\text{rel},i}$, or by root joint translation changes $\Delta t$, or even by any combination of the two. The non-zero off-diagonal elements $\partial u_i/\partial t_z$ and $\partial v_i/\partial t_z$ of the Jacobian matrix further indicate that depth estimation errors ``leak'' into the $u,v$ coordinate estimates in the image plane, and vice versa.

\subsection{Perspective Information Loss and Coupling Effect Analysis}
\label{subsec:app-perspective-loss-coupling}

Based on the analysis of the complete Jacobian matrix, this section explores the perspective information loss caused by cropping operations and its interaction with the perspective coupling effect. In practical applications, hand pose estimation typically first obtains a hand bounding box through a detector, then crops and scales the region within the bounding box as input to the network. This preprocessing step simplifies the input but introduces the problem of perspective information loss.

In the original image, the position and size of the hand contain rich perspective geometric information---the farther the hand is from the camera, the smaller the image size, and the projection process is closer to the orthogonal projection. From an information-theoretic perspective, cropping reduces the amount of information for depth estimation. In the original image, the hand position $(u_0, v_0)$ and size $s_0$ provide strong cues about depth $t_z$: according to perspective projection, size $s_0 \propto 1/t_z$. After cropping, these cues are partially removed, increasing the uncertainty of depth estimation.

The perspective information loss caused by cropping exacerbates the perspective coupling effect. After losing the original position information, the network can only infer global position from the content of the cropped image, and must rely on indirect cues in the image content, which are often ambiguous and unreliable.

From the perspective of the Jacobian matrix, perspective information loss caused by cropping is closely related to the following properties of the matrix:

\begin{enumerate}
    \item \textbf{Mixed influence of local pose and global position}: Matrix elements depend on both local pose $J_{\text{rel},i}$ and global translation $t$. For example, $\partial u_i/\partial t_z = -f_x X_i / Z_i^2$ contains both local pose component $J_{\text{rel},i,x}$ and global translation component $t_x$. This means that projection changes are simultaneously affected by local pose and root joint position, and the effects of the two on the image plane are inseparable.

    \item \textbf{Strong nonlinear coupling in the depth direction}: Depth changes $t_z$ not only affect projection through the denominator effect of $Z_i$ (near objects appear larger), but also produce cross-coupling through product terms with local coordinates $J_{\text{rel},i,x}, J_{\text{rel},i,y}$. When hand pose ($J_{\text{rel},i}$) changes, the same depth change $t_z$ produces different projection offsets.

    \item \textbf{Limitations of translation direction independence}: Horizontal translation $t_x$ only affects the $u$ coordinate, and vertical translation $t_y$ only affects the $v$ coordinate (diagonal elements of the matrix), but this independence is local because $t_x, t_y$ indirectly affect depth derivative terms by changing $X_i, Y_i$. The cropping operation loses position information in the original image, making it difficult for the network to exploit this limited independence to infer translation components.
\end{enumerate}

The non-zero off-diagonal elements $\partial u_i/\partial t_z$ and $\partial v_i/\partial t_z$ of the Jacobian matrix indicate that depth estimation errors ``leak'' into the $u,v$ coordinate estimates in the image plane, and vice versa. This coupling makes direct estimation of camera-space pose from monocular images an ill-posed problem: the same projection change $\Delta p_i$ may be caused by many different combinations of $(J_{\text{rel},i}, t)$.

\begin{table*}
  \centering
  \caption{Per-joint error change before and after root alignment on case of row \#1 of \cref{fig:ra-supp}. Negative $\Delta$ means the error decreases after alignment.}
  \label{tab:root_align_5_dex}
  \begin{tabular}{lrrr}
  \toprule
  Joint & Before & After & $\Delta$ (After - Before) \\
  \midrule
  Wrist    & 24.0174 &  0.0000 & -24.0174 \\
  Thumb\_1 & 15.9811 &  8.7894 &  -7.1917 \\
  Thumb\_2 & 13.5497 & 12.0409 &  -1.5088 \\
  Thumb\_3 & 12.5836 & 13.5603 &   0.9767 \\
  Thumb\_4 & 15.4669 & 14.2594 &  -1.2074 \\
  Index\_1 &  9.3494 & 20.2602 &  10.9108 \\
  Index\_2 &  8.5350 & 20.6546 &  12.1196 \\
  Index\_3 &  7.8767 & 19.4604 &  11.5837 \\
  Index\_4 &  8.9758 & 19.0349 &  10.0591 \\
  Middle\_1 &  9.5127 & 18.7216 &   9.2088 \\
  Middle\_2 &  5.8279 & 22.4163 &  16.5884 \\
  Middle\_3 &  2.7855 & 25.2093 &  \textbf{22.4238} \\
  Middle\_4 &  6.7389 & 29.1953 &  \textbf{22.4564} \\
  Ring\_1  & 11.7361 & 13.2491 &   1.5130 \\
  Ring\_2  &  7.3535 & 17.4583 &  10.1048 \\
  Ring\_3  &  4.3867 & 21.7554 &  17.3686 \\
  Ring\_4  &  9.7153 & 27.1253 &  \textbf{17.4100} \\
  Pinky\_1 & 15.2682 &  9.6775 &  -5.5907 \\
  Pinky\_2 & 13.6472 & 13.1983 &  -0.4490 \\
  Pinky\_3 & 14.1184 & 19.5819 &   5.4635 \\
  Pinky\_4 & 23.4038 & 30.9909 &   7.5871 \\
  \bottomrule
  \end{tabular}
\end{table*}

\begin{table*}
  \centering
  \caption{Per-joint error change before and after root alignment on case of row \#2 of \cref{fig:ra-supp}. Negative $\Delta$ means the error decreases after alignment.}
  \label{tab:root_align_9_dex}
  \begin{tabular}{lrrr}
  \toprule
  Joint & Before & After & $\Delta$ (After - Before) \\
  \midrule
  Wrist    & 20.9428 &  0.0000 & -20.9428 \\
  Thumb\_1 & 18.9970 &  1.9609 & -17.0360 \\
  Thumb\_2 & 15.5668 &  5.4748 & -10.0920 \\
  Thumb\_3 & 10.6571 & 10.4247 &  -0.2324 \\
  Thumb\_4 &  7.8074 & 15.1095 &   7.3021 \\
  Index\_1 & 10.1454 & 11.4248 &   1.2794 \\
  Index\_2 &  8.6302 & 20.2237 &  \textbf{11.5934} \\
  Index\_3 & 10.9266 & 22.7617 &  \textbf{11.8351} \\
  Index\_4 & 10.5297 & 21.4762 &  10.9464 \\
  Middle\_1 & 10.5208 & 12.7205 &   2.1997 \\
  Middle\_2 & 11.5187 & 15.0806 &   3.5619 \\
  Middle\_3 & 11.4452 & 14.9854 &   3.5403 \\
  Middle\_4 & 10.7450 & 13.7460 &   3.0009 \\
  Ring\_1  & 12.6135 & 11.9443 &  -0.6692 \\
  Ring\_2  & 11.9614 & 11.3974 &  -0.5640 \\
  Ring\_3  &  9.8892 & 11.2968 &   1.4076 \\
  Ring\_4  & 11.6317 & 16.0329 &   4.4013 \\
  Pinky\_1 & 14.5713 & 11.5837 &  -2.9876 \\
  Pinky\_2 & 12.3349 & 12.6961 &   0.3613 \\
  Pinky\_3 &  7.6028 & 14.3917 &   6.7889 \\
  Pinky\_4 &  3.4800 & 18.7052 &  \textbf{15.2252} \\
  \bottomrule
  \end{tabular}
\end{table*}

\begin{table*}
  \centering
  \caption{Per-joint error change before and after root alignment on case of row \#1 of \cref{fig:ra-ho3d-supp}. Negative $\Delta$ means the error decreases after alignment.}
  \label{tab:root_align_1_ho3d}
  \begin{tabular}{lrrr}
    \toprule
     Joint & Before & After & $\Delta$ (After - Before) \\
    \midrule
    Wrist & 19.4289 & 0.0000 & -19.4289 \\
    Thumb\_1 & 22.7313 & 4.1163 & -18.6150 \\
    Thumb\_2 & 17.3230 & 4.0984 & -13.2246 \\
    Thumb\_3 & 12.3879 & 7.0410 & -5.3469 \\
    Thumb\_4 & 12.5547 & 16.0392 & 3.4845 \\
    Index\_1 & 12.2225 & 9.3668 & -2.8557 \\
    Index\_2 & 5.3559 & 21.1668 & \textbf{15.8109} \\
    Index\_3 & 17.7139 & 32.0996 & \textbf{14.3857} \\
    Index\_4 & 43.3812 & 57.7669 & \textbf{14.3857} \\
    Middle\_1 & 8.0081 & 12.1308 & 4.1227 \\
    Middle\_2 & 8.1967 & 15.0336 & 6.8368 \\
    Middle\_3 & 11.5712 & 19.4476 & 7.8763 \\
    Middle\_4 & 15.8597 & 23.5558 & 7.6961 \\
    Ring\_1 & 7.3039 & 12.1250 & 4.8211 \\
    Ring\_2 & 6.5721 & 17.1794 & 10.6073 \\
    Ring\_3 & 10.6023 & 21.6348 & 11.0325 \\
    Ring\_4 & 11.6765 & 22.7090 & 11.0325 \\
    Pinky\_1 & 9.2459 & 15.0597 & 5.8139 \\
    Pinky\_2 & 17.6427 & 23.6320 & 5.9894 \\
    Pinky\_3 & 23.3074 & 29.2968 & 5.9894 \\
    Pinky\_4 & 29.5787 & 35.5681 & 5.9894 \\
    \bottomrule
  \end{tabular}
\end{table*}

\begin{table*}
  \centering
  \caption{Per-joint error change before and after root alignment on case of row \#2 of \cref{fig:ra-ho3d-supp}. Negative $\Delta$ means the error decreases after alignment.}
  \label{tab:root_align_2_ho3d}
  \begin{tabular}{lrrr}
    \toprule
    Joint & Before & After & $\Delta$ (After - Before) \\
    \midrule
    Wrist & 8.4425 & 0.0000 & -8.4425 \\
    Thumb\_1 & 10.2311 & 3.5969 & -6.6342 \\
    Thumb\_2 & 4.4493 & 8.8392 & 4.3899 \\
    Thumb\_3 & 21.4771 & 29.1681 & \textbf{7.6910} \\
    Thumb\_4 & 46.5958 & 54.2868 & \textbf{7.6910} \\
    Index\_1 & 21.6013 & 13.1588 & -8.4425 \\
    Index\_2 & 36.6657 & 28.2232 & -8.4425 \\
    Index\_3 & 38.3033 & 29.8608 & -8.4425 \\
    Index\_4 & 36.8067 & 28.3642 & -8.4425 \\
    Middle\_1 & 13.1708 & 4.7395 & -8.4313 \\
    Middle\_2 & 8.4589 & 6.5191 & -1.9398 \\
    Middle\_3 & 5.3740 & 12.8966 & \textbf{7.5227} \\
    Middle\_4 & 22.6355 & 30.3265 & \textbf{7.6910} \\
    Ring\_1 & 3.9240 & 5.6899 & 1.7659 \\
    Ring\_2 & 1.6087 & 8.9511 & 7.3424 \\
    Ring\_3 & 7.0678 & 14.7588 & \textbf{7.6910} \\
    Ring\_4 & 18.9014 & 26.5924 & \textbf{7.6910} \\
    Pinky\_1 & 3.9506 & 9.6210 & 5.6703 \\
    Pinky\_2 & 5.8738 & 9.6193 & 3.7455 \\
    Pinky\_3 & 10.7508 & 12.5462 & 1.7953 \\
    Pinky\_4 & 12.6621 & 19.2628 & 6.6007 \\
    \bottomrule
  \end{tabular}
\end{table*}


\end{document}